\documentclass[twoside]{article}

\usepackage[accepted]{arxiv}

\usepackage[round,numbers]{natbib}

\setcitestyle{authoryear}
\usepackage{mathtools} 

\usepackage{amsmath}
\usepackage{amssymb}
\usepackage{mathtools}
\usepackage{amsthm}
\usepackage{dsfont}

\newcommand{\ie}{\textit{i}.\textit{e}., }
\newcommand{\eg}{\textit{e}.\textit{g}., }

\usepackage[utf8]{inputenc} %
\usepackage[T1]{fontenc}    %
\usepackage{hyperref}       %
\usepackage{url}            %
\usepackage{booktabs}       %
\usepackage{amsfonts}       %
\usepackage{nicefrac}       %
\usepackage{microtype}      %
\usepackage{xcolor}         %

\usepackage{graphicx}
\usepackage{subcaption}

\usepackage{algorithm}
\usepackage{algpseudocode}

\newtheorem{lemma}{Lemma}
\newtheorem{theorem}{Theorem}
\newtheorem{assumption}{Assumption}

\providecommand{\regret}{\textsf{Regret}}
\providecommand{\checkx}{\check{x}}

\providecommand{\UCB}{\textsf{UCB}}
\providecommand{\LCB}{\textsf{LCB}}
\providecommand{\CI}{\textsf{CI}}

\providecommand{\HRbandit}{\textsf{HR-Bandit}}

\usepackage[capitalize,noabbrev]{cleveref}
\usepackage{abstract}
\usepackage{authblk}
\begin{document}

\twocolumn[
\aistatstitle{HR-Bandit: Human-AI Collaborated Linear Recourse Bandit}

\aistatsauthor{Junyu Cao\footnotemark[1] \And Ruijiang Gao\footnotemark[1] \And  Esmaeil Keyvanshokooh\footnotemark[1]} 

\aistatsaddress{University of Texas at Austin\And  University of Texas at Dallas \And 
Texas A\&M University} ]
\footnotetext[1]{Alphabetical Order.}

\begin{abstract}
Human doctors frequently recommend actionable recourses that allow patients to modify their conditions to access more effective treatments. Inspired by such healthcare scenarios, we propose the Recourse Linear UCB (\textsf{RLinUCB}) algorithm, which optimizes both action selection and feature modifications by balancing exploration and exploitation. We further extend this to the Human-AI Linear Recourse Bandit (\textsf{HR-Bandit}), which integrates human expertise to enhance performance. \textsf{HR-Bandit} offers three key guarantees: (i) a warm-start guarantee for improved initial performance, (ii) a human-effort guarantee to minimize required human interactions, and (iii) a robustness guarantee that ensures sublinear regret even when human decisions are suboptimal. Empirical results, including a healthcare case study, validate its superior performance against existing benchmarks.
\end{abstract}

\section{Introduction}

Consider a clinical trial setting in which physicians seek optimal personalized treatments for a chronic condition, such as hypertension, based on the latest patient information. Such information includes the patient's medical history, lifestyle, and most recent lab results, all of which serve as the {\em context} for the patient's treatment. Given no prior knowledge about the disease and some available treatment plans, physicians may resort to contextual bandit algorithms such as Linear Upper Confidence Bound (LinUCB) \citep{chu2011contextual} to decide which personalized treatment plan should be delivered to each patient adaptively. However, physicians notice that the best treatment  options may not be as effective for the patient given their current features. Instead, they advise the patient to modify certain mutable features by providing them with {\em recourses}, such as lowering the blood sugar level to 120 mg/dL \citep{american201910}, with the aim of qualifying for a more effective treatment option in the future. ~\looseness=-1

The practical concept of recourse proposed by physicians parallels the idea of algorithmic recourse, or counterfactual explanations, which has gained significant attention in the machine learning community, particularly in the context of {\em offline classification}, where the classifier remains fixed \citep{wachter2017counterfactual,verma2020counterfactual}. These works aim to   assist individuals with unfavorable AI decisions by providing algorithmic recourse. In online learning, however, the uncertainty in the true decision function introduces additional complexity in offering recourses. Here, recourse is understood as a means for patients to modify their features to achieve a more favorable outcome. Motivated by this research gap, we propose \textsf{\underline{R}ecourse \underline{Lin}ear \underline{U}pper \underline{C}onfidence \underline{B}ound} (\textsf{RLinUCB}) that synergizes the strengths of (i)  contextual bandit algorithms with (ii) algorithmic recourse in an {\em online} learning setting.

Given a patient with immutable features (\eg age) and mutable features (\eg blood sugar level, lifestyle), \textsf{RLinUCB} aims to jointly optimize for both the action (treatments) and context (mutable patient features). Specifically, our recourse bandit approach helps physicians achieve a two-fold objective: (i) prescribing a proper personalized treatment plan, and (ii) suggesting modifications to the current mutable features, aiming to maximize the patient's health benefits. By doing so, \textsf{RLinUCB} helps the patient to strengthen their conditions by offering {\em counterfactual explanations} so that the patient can improve their treatment outcome or qualify for a more effective treatment plan. 
In addition, as more patients arrive, \textsf{RLinUCB} can learn  adaptively and refine its recommendations.

\begin{figure*}[h]
  \centering
    \includegraphics[width=\textwidth]{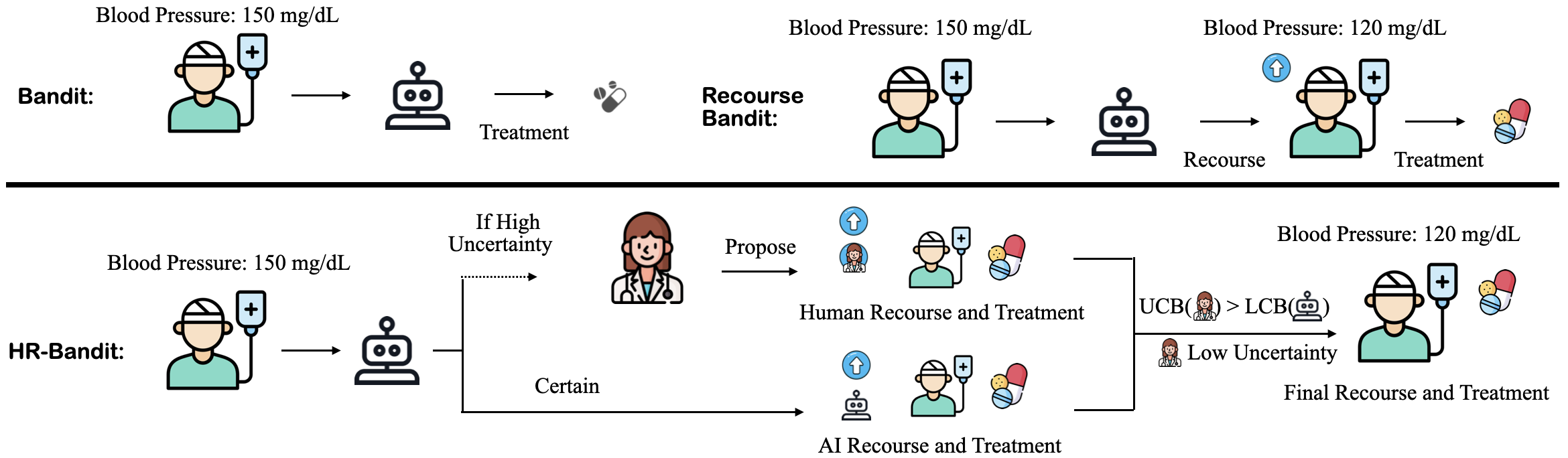}
    \caption{Illustration of \textsf{Recourse Bandit} and \textsf{HR-Bandit}. \textsf{Recourse Bandit} offers algorithmic recourses to patients to improve their health conditions for more effective treatment.  \textsf{HR-Bandit} selectively consults human experts based on uncertainty estimates, enabling a data-driven integration of human expertise and AI recommendations.~\looseness=-1}
    \label{fig:figure1}
\end{figure*}

However, like other bandit algorithms, \textsf{RLinUCB} must learn from scratch through exploration, which may result in suboptimal recourses and treatments for early patients, even though physicians already possess (albeit imperfect) domain knowledge about recourse and treatment recommendations. While there is extensive research on human-AI collaboration in offline settings \citep{madras2018predict,wilder2020learning,raghu2019algorithmic}, leveraging human expertise to warm-start bandit algorithms remains a largely under-explored area. In light of this, we propose \textsf{\underline{H}uman-AI Linear \underline{R}ecourse \underline{Bandits}} (\textsf{HR-Bandit}) that selectively consults human experts to improve patient outcomes. Specifically, \textsf{HR-Bandit} first assesses the AI's uncertainty regarding the recourse recommendation: if the AI is confident, it autonomously prescribes the recourse; if the uncertainty is high, a human expert is consulted for a recommendation. A data-driven selection criterion is then used to determine the appropriate decision-maker. A detailed illustration of \textsf{RLinUCB} and \textsf{HR-Bandit} is provided in \Cref{fig:figure1}.

\textsf{HR-Bandit} offers several advantages as a human-AI system. First, it makes no assumptions about the human expert's behavior model, requiring only black-box queries from humans. Additionally, \textsf{HR-Bandit} benefits from high-quality human input, remains robust to erroneous queries, and minimizes the need for human interventions. These advantages are formally defined as: ~\looseness=-1

\begin{itemize}
    \item \textbf{Warm-start guarantee}: We show that \textsf{HR-Bandit} reduces regret when human experts perform well. Intuitively, human experts can guide the bandit algorithm toward better decisions early on. ~\looseness=-1
    \item \textbf{Robustness guarantee}: We demonstrate that our algorithm achieves the same convergence rate in the worst-case regret as a standard bandit algorithm without human experts, even when decision quality is low. This robustness is ensured through data-driven control of the selection criteria. 
    \item \textbf{Human effort guarantee}: In an ideal human-AI system, human interventions are minimized. We ensure that human interactions in the system are required only for a finite number of times, leading to a more automated human-AI system. ~\looseness=-1
\end{itemize}

We summarize our main contributions as follows:

\begin{enumerate}
    \item We introduce \textsf{Recourse Bandit} problem and propose \textsf{RLinUCB} that jointly determines the minimal feature modifications over the mutable features and provides action recommendations. We show theoretically \textsf{RLinUCB} has a $\tilde{O}(d\sqrt{KT})$ regret. 
    \item We develop \textsf{HR-Bandit} that extends \textsf{RLinUCB} into a human-AI collaboration framework that leverages human domain knowledge for guiding the AI algorithm. We derive the recourse regret of $\tilde{O}(\min\{\eta T, d\sqrt{KT}\})$ for this algorithm, where $\eta$ reflects the quality of human's decision.
    \item We show theoretically \textsf{HR-Bandit} has 1) {\em warm-start guarantee} that reduces initial regret in the early rounds with the help of human experts; 2) {\em robustness guarantee} that an {\em adversarial} human expert will not impact the convergence rate; and 3) {\em human effort guarantee} that \textsf{HR-Bandit} only requires a limited number of human interventions. 
    \item Using both synthetic data and real-world healthcare datasets, we empirically demonstrate that our \textsf{RLinUCB} and \textsf{HR-Bandit} algorithms outperform the standard linear contextual bandit algorithm and empirically validate their theoretical properties. ~\looseness=-1
\end{enumerate}

\section{Related Work}

\noindent\textbf{Algorithmic Recourse}: 
Algorithmic Recourse seeks to help people who have been negatively impacted by algorithmic decisions by offering alternative paths to positive outcomes~\citep{wachter2017counterfactual,ustun2019actionable,van2019interpretable,pawelczyk2020learning,mahajan2019preserving,karimi2019model,karimi2020probabilistic,dandl2020multi,gao2023impact}.  These methods can be classified in several ways~\citep{verma2020counterfactual}, including the type of predictive model used, the level of access to the model, the preference for changing only a few features in the suggested alternatives (sparsity), whether the alternatives should resemble real data, the consideration of causal relationships, and whether the method produces one or multiple alternatives.  Research has also shown that current recourse methods may lack robustness, as small changes to the initial input~\citep{dominguezolmedo2021adversarial} or the underlying model~\citep{upadhyay2021robust,rawal2021modelshifts} can affect the suggested recourse.  Most prior research in algorithmic recourse focuses on offline classification.  This work introduces the first algorithmic recourse approach within a contextual bandit framework, combining recourse with online learning to potentially improve the performance of online algorithms by providing counterfactual explanations.

\noindent\textbf{Contextual Bandits}: Contextual bandits have been surveyed by \citet{slivkins2019introduction, bubeck2012regret, lattimore2020bandit}. There is indeed a vast and growing body of literature on developing various contextual MAB algorithms for different settings. \citet{auer2002using} developed the first algorithm for the linear contextual MAB. Several works then proposed both UCB-based (\eg \citep{dani2008stochastic, chu2011contextual, abbasi2011improved, li2017provably,gan2024contextual}) and Thompson sampling-based (\eg \citep{agrawal2013thompson, keyvanshokooh2025contextual, russo2014learning, abeille2017linear}) algorithms for contextual MABs. \citet{harris2022strategy} consider the case where human decision-makers act adversarially to a known public policy and prove that deriving a sublinear strategic regret is {\em generally impossible}. Unlike previous work, we consider the unique \textit{recourse bandit} setting where human subjects try to improve their conditions (features) by trying to follow the algorithm's recommendations and theoretically prove that our algorithm achieves a sublinear \textit{recourse regret}. 
Our work is also closely related to the hint-based bandit \citep{cutkosky2022leveraging,bhaskara2023bandit}. Unlike these methods that only have a single hint in the beginning, in recourse bandit, we can receive a different context at each time step and need to decide whether to consult human experts for a different recourse.

Our work is related to conversational bandits \citep{zhang2020conversational, zuo2022hierarchical}. 
Conversational bandits focus on incorporating conversations through asking questions (e.g., user preferences on key-terms) to accelerate learning. The main objective of their setup is to learn user preferences and reduce regret faster by leveraging extra conversational feedback beyond typical contextual bandit feedback. This setup is different from HR-bandit, where we observe human recommendations and select decisions from either the human or the AI through uncertainty comparisons.

\noindent\textbf{Human-AI Collaboration}: Human-AI collaboration has been studied extensively in the offline classification setting \citep{madras2018predict,wilder2020learning,raghu2019algorithmic,gao2024sel}. A standard human-AI collaboration form is the decision support design where humans have access to AI recommendations to make decisions \citep{bansal2019case,de2018learning,boyaci2024human}.
Recent works study the human-AI collaborated in the offline reinforcement learning (RL), \citet{bastani2021improving,grand2024best} use RL generated actions to guide human decision making. \citet{gao2021human,gao2023confounding} design learning-to-defer framework with bandit feedback to allocate tasks between humans and AI. 
\citet{bordt2022bandit,wang2022blessing} study how private information of humans can impact and be incorporated in RL and bandit algorithms. 
Unlike these works, we focus on improving online bandit algorithms using experts' domain knowledge and show humans can improve bandit algorithms even with the same information.

\section{Linear Recourse Bandits}\label{sec: LCRB}

\subsection{Model}
Consider the following decision-making problem with contextual information.  Prior to decision-making,
the decision-maker (DM) observes a context $x=(x_{I}, x_{M})\in \mathcal{X}_I \times \mathcal{X}_M$ where $x_{I}$ is the {\em immutable} context such as age and sex, and $x_{M}$ is the {\em mutable} context such as blood sugar level and average alcohol consumed, and $\mathcal{X}_I$ and $\mathcal{X}_M$ are feasible regions of immutable and mutable contexts, respectively \citep{ustun2019actionable}.  Given this context, the conditional expected reward of a decision/action $A = a$, where $a\in \mathcal{A}$, is parameterized by $\theta_a^*=(\theta_{a,I}^*, \theta_{a,M}^*)$ as the following linear model:
\[
\begin{aligned}
r(x,a):=& E[R|X=x, A=a] = \theta_a^*{}^\top x\\
= & x_I^\top\theta_{a,I}^* +  x_M^\top \theta_{a,M}^*.
\end{aligned}
\]

The action set, which can be different treatment plans, for example, contains $K$ different actions in total, \ie $|\mathcal{A}|=K$. The immutable context cannot be changed, while we allow the mutable context to be changed around $x_M$ as long as the distance is controlled. More specifically, we allow $x_M$ to be changed to $\check{x}_M$ as long as $d(\checkx_M, x_M) \leq \gamma$, where $d(\cdot, \cdot)$ is some distance function and $\gamma$ is a constant. For example, we can choose the two-norm or the infinite-norm distance, \ie $d(\checkx_M, x_M)=\|\checkx_M -x_M\|_2$ or $d(\checkx_M, x_M)=\|\checkx_M -x_M\|_{\infty}$. When the distance function is the two-norm distance, the features are considered independent and can be modified individually. Correlated features can correspond to more general distance functions such as the mahalanobis distance. Moreover, future work can consider causal recourse \citep{von2022fairness} that features follow a given structural causal model.

Given a context $x = (x_{I}, x_{M})$ and the true model parameter $\{\theta_{a}^*\}_{a\in \mathcal{A}}$, we select the best action {\em and} ask the mutable context $x_M$ to be changed to $\checkx_M$ using the new {\em counterfactual optimization problem} (COP): 
\begin{align}
    \max_{a\in \mathcal{A}}\max_{\checkx_M\in \mathcal{X}_M} &\quad  (x_I,\checkx_M)^\top \theta_a^*\tag{\textsf{COP}}\label{E. optimization}\\
    \quad s.t. &\quad d(\checkx_M, x_M) \leq \gamma. \notag
\end{align}

In \eqref{E. optimization}, the objective is to maximize the expected reward such that the modified context is within $\gamma$-distance away from $x_M$. The optimal solution depends on the selection of the distance function. Lemma \ref{lemma: optimal solution} provides a closed-form solution when the distance is chosen as the two-norm.

\begin{lemma}[\textbf{Closed-form Solution for Two-norm Distance}]\label{lemma: optimal solution}
    Let $d(x,x') = \|x-x'\|_2$ be the distance function, then the optimal solution to \eqref{E. optimization} is $\checkx_M^*=x_M+\theta_{a,M}^*/\|\theta_{a,M}^*\|\gamma$ for a given action $a$.
\end{lemma}

We prove Lemma \ref{lemma: optimal solution} by showing the KKT point of \eqref{E. optimization}. In the main body of this paper, we either omit full proofs or provide only proof sketches. Complete proofs can be found in Appendix \ref{app:proof}.

When the distance is controlled along each dimension, then the optimization problem \eqref{E. optimization} is separable, and we can optimize for each dimension in sequence. ~\looseness=-1

\begin{lemma}[\textbf{Closed-form Solution for Box Constraint}]\label{lemma: optimal solution2}
    When the mutable context is constrained by $|x_M(j)-\checkx_M(j)|\leq \gamma_j$ for all $j$ corresponding to the mutable dimension, then the optimal solution to \eqref{E. optimization} is $\checkx_M^*(j)= x_M(j)+\theta_{a,M}^*(j)/|\theta_{a,M}^*(j)| \gamma_j$.
\end{lemma}

\subsection{Optimistic Counterfactual Optimization}
In the previous section, we discussed the counterfactual optimization problem when the true model parameter is known. However, in real-world scenarios, the unknown parameter $\theta_{\cdot}^*$ must be learned via an online decision-making process.

Given historical observations $\{(y_s, x_s, a_s)\}_{s=1}^{t-1}$, suppose that the unknown model parameter $\theta_{a}^*$  can be estimated by minimizing the following squared-loss problem:
\[
\hat{\theta}_{ta} \in \arg\min_{\theta=(\theta_{a,M}, \theta_{a,I})\in \mathbb{R}^d} \  \sum_{s=1}^{t-1} (y_s-\theta^\top x_s)^2 \; \mathds{1}(a_s= a).
\]

In the linear bandits literature, it is well-known that the uncertainty set for the  model parameter $\theta_{a}^*$ can be constructed by $\Theta_{ta}:=\{\theta_a:\|\theta_a-\hat{\theta}_{ta}\|_{V_{ta}}\leq \rho_{ta}\}$ (see Lemma \ref{lemma: radius}  in Appendix \ref{appendix: proof}), 
where the design matrix is
$V_{ta}:= \sum_{s=1}^{t-1} x_s \, x_s^\top \mathds{1}(a_s=a) + I$, and the radius $\rho_{ta}=\beta_{\Theta}+\sqrt{2\log\left(\frac{K}{\delta}\right)+d\log\left(1+\frac{\sum_{s=1}^{t-1} \mathds{1}(a_t=a) \beta_\mathcal{X}}{d}\right)}$, which is chosen according to Lemma \ref{lemma: radius}. Having constructed this uncertainty set $\Theta_{ta}$, we solve an {\em optimistic} version of the counterfactual optimization problem \eqref{E. optimization}:
\begin{align}
\label{E. optimization2}
   \max_{a\in \mathcal{A}} \max_{\checkx_M\in \mathcal{X}_M,\theta_a\in \Theta_{ta}} &\quad \checkx_M^\top \theta_{a,M} +x_I^\top \theta_{a,I}\tag{\textsf{OCOP}}\\
    \quad  s.t. &\quad d(\checkx_M, x_M) \leq \gamma. \notag
    \end{align}
   
In \emph{optimistic counterfactual optimization problem} \eqref{E. optimization2}, it maximizes the expected reward when $\checkx_M$ is within $\gamma$-distance from the initial mutable context $x_M$ and $\theta_a$ is contained in the uncertainty set $\Theta_{ta}$. When the true parameter $\theta_a^*$ is contained in the uncertainty set, the optimal solution of \eqref{E. optimization2} is an {\em upper bound} of \eqref{E. optimization}. ~\looseness=-1

\textbf{Solving the Optimistic Problem \eqref{E. optimization2}$\;\;$}
We note that the optimistic counterfactual optimization problem \eqref{E. optimization2} is a non-linear constrained optimization problem. To solve it efficiently, we deploy the alternating direction method of multipliers (ADMM). This augmented Lagrangian-based method is a common approach for solving complex optimization problems \citep{boyd2011distributed, han2022survey}. 
Now, let $\lambda_{\delta} \in \mathbb{R}
$ and $\lambda_{\rho} \in \mathbb{R}$ be the Lagrangian multipliers corresponding to each constraint of \eqref{E. optimization2}, respectively, and also let $\beta_{\gamma} > 0
$ and $\beta_{\rho}$ 
be penalty parameters. Then, we can formulate the augmented Lagrangian function associated with \eqref{E. optimization2} as follows:

{\small 
\begin{align*}\label{eq: aug-lagrangian}
    &\mathcal{L}_{\beta_{\gamma}, \beta_{\rho}}(\checkx_M, \theta_{a}, \lambda_{\gamma}, \lambda_{\rho}) \\
    =& \; (x_I,\checkx_M)^\top \theta_a + \lambda_{\gamma} \max\{ d(\checkx_M, x_M) - \gamma, 0 \}^2 \\
    &- \dfrac{\beta_{\gamma}}{2} \Big( \max\{ d(\checkx_M, x_M) - \gamma, 0 \}^2 \Big)^2
     \\
    &+ \lambda_{\rho} \max\{ \|\theta_{a} -\hat{\theta}_{t a}\|_{V_{ta}} - \rho_{ta}, 0 \}^2\\
    &
    - \dfrac{\beta_{\rho}}{2} \left(  \max\{ \|\theta_{a} -\hat{\theta}_{t a}\|_{V_{ta}} - \rho_{ta}, 0 \}^2 \right)^2.
\end{align*}
}

Accordingly, we can develop the following iterative scheme of the ADMM approach for solving the counterfactual optimization problem \eqref{E. optimization2} for each action $\mathcal{A}$ in Algorithm \ref{alg:ADMM}. Then we select the best action with the highest optimistic reward.
\begin{algorithm}
\caption{ADMM Mechanism for Solving the Optimistic Counterfactual Optimization  \eqref{E. optimization2}}\label{alg:ADMM}
\begin{algorithmic}
\State \textbf{Input:} A counterfactual 
$\checkx_M = \checkx_M^{(0)}$,
the parameters $\theta_a = \theta_a^{(0)}$, $\lambda_{\gamma} = \lambda_{\gamma}^{(0)}$, 
 and $\lambda_{\rho} = \lambda_{\rho}^{(0)}$.  
\State \textbf{Repeat for $k = 1, 2, ...$ as:}
\begin{align}
& \checkx_M^{(k+1)} = \arg\max_{\checkx_M}  \mathcal{L}_{\beta_{\delta}, \beta_{\rho}}(\checkx_M, \theta_{a}^{(k)}, \lambda_{\delta}^{(k)}, \lambda_{\rho}^{(k)})
\tag{\textrm{\emph{Optimizing $\checkx_M$}}} \\
& \theta_{a}^{(k+1)} = \arg\max_{\theta_{a}}  \mathcal{L}_{\beta_{\delta}, \beta_{\rho}}(\checkx_M^{(k)}, \theta_{a}, \lambda_{\delta}^{(k)}, \lambda_{\rho}^{(k)})
\tag{\textrm{\emph{Optimizing $\theta_{a}$}}}\;
\\
& \lambda_{\delta}^{(k+1)} =  \lambda_{\delta}^{(k)} - \beta_{\delta} \Big( \max\{ d(\checkx_M^{(k+1)}, x_M) - \gamma, 0 \}^2\Big) \tag{\textrm{\emph{Gradient Ascent on $\lambda_{\delta}$}}} \\
& \lambda_{\rho}^{(k+1)} = \lambda_{\rho}^{(k)} - \beta_{\rho} \left( \max\{ \|\theta_{a}^{(k+1)} -\hat{\theta}_{t a}\|_{V_{ta}} - \rho_{ta}, 0 \}^2 \right)
\tag{\textrm{\emph{Gradient Ascent on $\lambda_{\rho}$}}} 
\end{align}
\end{algorithmic}
\end{algorithm}

We note that there are theoretical works proving the convergence of the ADMM mechanism under various assumptions (\eg \citep{hong2017linear, wang2019global, hong2016convergence}). In our \eqref{E. optimization2}, we have a nonconvex objective function and convex constraints. For this setting, the convergence of our proposed ADMM Algorithm \ref{alg:ADMM} is guaranteed following the work of \citep{ wang2019global}.

It is worth noting that our ADMM algorithm is flexible to include other practical constraints desired for algorithmic recourses \citep{verma2020counterfactual}. For example, we may want to put a constraint on the number of mutable features (i.e., sparsity constraints) to reduce human effort \citep{miller2019explanation}; that is, $\sum_{i=1}^m \mathds{1}(\checkx_M[i] \neq x_M[i]) \leq u$, where $u$ is the sparsity level. 
For the other example, we may want to change the context to a ``realistic'' one, meaning the recourse adheres to the underlying distribution \citep{pawelczyk2020learning,joshi2019towards}. To be more specific, let's say there is a known probability density function over the context space $\mathcal{X}$, denoted as $p(x)$. The suggested modifications should satisfy that $p(\checkx)>\underline{p}$ for some $\underline{p}>0$. 

\subsection{RLinUCB and Regret Analysis}
In this section, we analyze the  performance of our  
Recourse Linear Upper Confidence Bound (\textsf{RLinUCB}) algorithm (Algorithm \ref{alg:cap}).
We use the term {\em recourse regret} to measure the performance of our learning algorithm. For a policy $\pi\in\Pi$, it is defined as
   \[
   \regret_\pi(T) = E_\pi\left[\sum_{t=1}^T  \  r(x_t^*, a_t^*) - r(x_t, a_t) \right],
   \]
where $x_{t}^*=(x_{tI}^*, \checkx_{tM}^*)$ is the optimal recourse and $a_t^*$ is the optimal action at time $t$. It measures the difference between our decision and the optimal action. To analyze the regret, we show that the objective value computed by \eqref{E. optimization2} is optimistic in the face of uncertainty. Then we show a small perturbation around the optimal solution would only incur a small difference. By combining these two results, we quantify the one-step regret. In the part of regret analysis, we assume that Algorithm \ref{alg:ADMM} can find the optimal solution.

\begin{algorithm}
\caption{\textsf{RLinUCB} Algorithm}\label{alg:cap}
\begin{algorithmic}
\State \textbf{Input} Time horizon $T$.\;
\For{$t=1,\cdots,T$}
\State Solve the optimistic counterfactual optimization problem \eqref{E. optimization2} by Algorithm \ref{alg:ADMM}.\; %
\State Recommend the recourse $x_{t}$ and the action $a_t$.\;
\EndFor
\end{algorithmic}
\end{algorithm}

\begin{assumption}[Boundedness]\label{assump: bound}
Assume
$\underline{\beta}_{\mathcal{X}}\leq\|x\|\leq \beta_{\mathcal{X}}$ for all $x\in \mathcal{X}$ and $\sup_{\theta\in \Theta} \|\theta\|\leq \beta_{\Theta}$.
\end{assumption}

\begin{theorem}[\textbf{Regret of \textsf{RLinUCB}}]\label{thm: regret}
With probability at least $1-\delta$, the regret of Algorithm \ref{alg:cap} satisfies
\[
\regret_T\leq  \rho_T \sqrt{2 d KT\log\left((d+T\beta_\mathcal{X}^2)/d\right)}\,,
\]
where $\rho_T= \beta_{\Theta}+\sqrt{2\log(\frac{K}{\delta})+d\log(1+\frac{T \beta_\mathcal{X}}{d})}$.
\end{theorem}
\noindent \textbf{Proof Sketch:} Suppose $x_t=(x_I,\checkx_M)$ and $\theta_{ta}$ is the optimal solution to \eqref{E. optimization2}, then the one-step regret can be bounded by 
\[
r(x_t^*, a_t^*) - r(x_t, a_t) \leq \theta_{ta^*_t}^\top x_t- \theta_{ta}^\top x_t\leq\rho_{ta} \|x_t\|_{V_{ta}^{-1}}.
\]
By summing up the one-step regret until time step $T$ and applying Elliptical Potential Lemma, we are able to bound the total recourse regret. The detailed proof is included in Appendix \ref{appendix: proof}.\hfill$\Box$

Theorem \ref{thm: regret} characterizes the regret as $\tilde{O}(d\sqrt{KT})$. The result matches the regret lower bound for the linear bandits problem. It indicates that, though the decision is more complicated as it needs to simultaneously select the action and suggest context modifications, no additional regret is incurred based on our proposed algorithm. ~\looseness=-1

\section{Human-AI Linear Recourse Bandits}
In this section, we propose a new algorithm which we call \textsf{Human-AI Linear Recourse Bandits} (\textsf{HR-Bandit}). We will analyze the performance of \textsf{HR-Bandit} and discuss its improvements. \textsf{HR-Bandit} handles three technical challenges for human-AI systems. Specifically, when the human performance declines, the AI algorithm remains robust and does not fail, creating a reliable human-AI system ({\em robustness guarantee}). Conversely, when the human does perform well, the human domain knowledge needs to be leveraged to efficiently warm-up the AI algorithm, resulting in a more effective human-AI system than an AI-only approach ({\em warm-start guarantee}). Additionally, we ensure that human interaction and feedback to the system are required only for a finite number of times, leading to a more automated human-AI system ({\em human-effort guarantee}). Our \textsf{HR-Bandit} algorithm provides all these three types of performance guarantees (formally presented in \S \ref{performances}).

\subsection{HR-Bandit Algorithm}
For any context vector $x$, we define the upper confidence bound of the context $x$ and action $a$ as $\UCB_t(x,a)= x^\top \hat{\theta}_{ta}+\rho_{ta}\|x\|_{V_{ta}^{-1}}$,  $\LCB_t(x,a)=x^\top \hat{\theta}_{ta}-\rho_{ta}\|x\|_{V_{ta}^{-1}}$, and $\CI_t(x,a)=\rho_{ta}\|x\|_{V_{ta}^{-1}}$ for notation simplicity.

The main steps of our proposed \HRbandit\ algorithm are as follows. It first generates the \UCB\  recourse $a^U$ and $\checkx^U_M$ by solving the optimistic problem \eqref{E. optimization2} at each iteration $t$. The algorithm then asks for the human recourse  $\checkx^H_M$ (which could be adversarial) if the generated  \UCB\  recourse $a^U$ and $\checkx^U_M$ are not reliable enough, i.e., $\UCB(x^U,a^U)-\LCB(x^U,a^U)>\Delta$, where $\Delta$ is a threshold parameter set by the decision maker. Once the algorithm has the information on both the human and \UCB\ decisions, it then decides which one to implement. Specifically, if (i) the upper confidence bound $\UCB(x^H,a^H)$ of the human decision $(x^H,a^H)$ is at least better than the lower confidence bound $\LCB(x^U,a^U)$ of the \UCB\ decision $(x^U,a^U)$, and also (ii) the information variance control condition $\CI_t(x^U, a^U)<\zeta \CI_t(x^H, a^H)$ holds, the human recourse and action then will be implemented. Otherwise, the algorithm will perform the \UCB\ recourse and action. Once such decisions are implemented, then we receive the reward and update the confidence set. The details of our proposed \HRbandit\ algorithm are provided in Algorithm \ref{alg: multiple action}.

\begin{algorithm}
\caption{\textsf{HR-Bandit} Algorithm}\label{alg: multiple action}
\begin{algorithmic}
\For{$t\in[T]$}
\State Generate the \UCB\  recourse $\checkx^U_M$ and action $a_t^U$. Let $x_t^U=(x^U_{tI}, \check{x}_{tM}^U)$.
\If{$\UCB_t(x_t^U,a_t^U)-\LCB_t(x_t^U,a_t^U)>\Delta$}
\State Human proposes recourse $x_t^H$ and $a_t^H$. 
\EndIf
\If{$\CI_t(x_t^U, a_t^U)<\zeta \CI_t(x_t^H, a_t^H)$, $\UCB_t(x_t^H,a_t^H)>\LCB_t(x_t^U,a_t^U)$, 
and also human recourse is generated} 
\State Set the recourse $x_t = x_t^H$ and select $a_t=a_t^H$
\Else
\State Set the recourse $x_t = x_t^U$ and select $a_t=a_t^U$
\EndIf 
\State Implement the recourse $x_t$ and the selected $a_t$; receive a corresponding reward as the feedback. 
\State Update the uncertainty set $\Theta_{ta}$. 
\EndFor 
\end{algorithmic}
\end{algorithm}

\subsection{Regret Analysis}\label{performances}
First, we show the warm-start guarantee. If implementing Algorithm \ref{alg:cap}, then the one-step regret is $2\CI(x^U, a^U)$. When the quality of the human's action is high, then $\textsf{HR-Bandit}$ can achieve a significant improvement, especially in the initial rounds. We show Theorem \ref{thm: warm start multiple actions} under the condition that the true reward falls into the range of $\LCB$ and $\UCB$, which holds with high probability. 
All proofs are included in \Cref{app:proof}. 
~\looseness=-1

\begin{theorem}[Warm-Start Guarantee]\label{thm: warm start multiple actions}
    Suppose there exists a constant $\eta>0$ such that $r(x^H,a^H)\geq r(x^*,a^*)-\eta$. Moreover, assume there exists a constant $\zeta>0$ such that for any $(x_1, a_1), (x_2, a_2)\in\mathcal{X}$, it holds that $\CI(x_1, a_1)<\zeta \CI(x_2,a_2)$. By setting $\Delta=\eta$, then the one-step regret can be bounded by ~\looseness=-1
     \[
    \begin{aligned}
  2\min\{\eta, (1+\zeta)\CI(x^U,a^U)\}.
    \end{aligned}
    \]   
\end{theorem}
\noindent \textbf{Proof Sketch:} To prove Theorem \ref{thm: warm start multiple actions}, we analyze two distinct scenarios: (1) When $\UCB(x^U,a^U)-\LCB(x^U,a^U)>\Delta$---the confidence interval is not narrow enough; and (2) When $\UCB(x^U,a^U)-\LCB(x^U,a^U)\leq\Delta$---the confidence interval is narrow enough. In scenario (1), if human's action is taken,  this implies there is some overlap between the confidence interval of the human's action and the AI's upper confidence bound solution, suggesting the human's action is reasonably good.  In this case, the regret can be bounded by the minimum of $2\eta$ and $2(1+\zeta)\CI(x^U, a^U)$. In scenario (2), if the confidence interval of AI solution is narrow enough, then the AI action will also be good enough.
\hfill$\Box$

The theorem above demonstrates the improvement guarantee of the \textsf{HR-Bandit} algorithm, especially during the early rounds and when human actions are of relatively high quality. We will discuss the implications of the assumption in Theorem \ref{thm: warm start multiple actions} in \Cref{app:assumption} due to space constraint. ~\looseness=-1

The next theorem establishes the human-effort guarantee, which ensures that human interaction with the system is required only a finite number of times under certain regularity conditions.

\begin{theorem}[Human-Effort Guarantee]\label{thm: human effort}
Suppose $\lambda_{\min} (V_{ta})\geq \omega n_{ta}^\beta$ for all $a\in \mathcal{A}$ where $\omega$, $\beta>0$, and $n_{ta}=\sum_{s=1}^{t-1} 1(a_t=a)$. For fixed confidence level $1-\delta$, the maximum time that human's proposal will be taken is $Kn^\diamond$ where 
\[
n^{\diamond}:= \max\left\{\left(\frac{6\beta_{\Theta}\beta_{\mathcal{X}}}{\Delta \sqrt{\omega}}\right)^{\frac{2}{\beta}}, \left(\frac{12\log(K/\delta)\beta_{\mathcal{X}}}{\Delta\sqrt{\omega}}\right)^{\frac{2}{\beta}}, \bar{n}\right\}
\]
and
$\bar{n}=\max\{n': \log(1+\frac{\beta_{\mathcal{X}}}{d} n')\leq  \frac{\Delta \sqrt{\omega} n'{}^{\beta/2}}{6d \beta_{\mathcal{X}}}\}$.
\end{theorem}
\noindent \textbf{Proof Sketch}: Algorithm \ref{alg: multiple action} specifies that no further recourse is proposed by the human when the condition $\UCB_t(x^U,a^U) - \LCB_t(x^U,a^U) \leq \Delta$ is met. This condition is satisfied when the confidence interval narrows as the number of pulls increases. Consequently, once the number of pulls exceeds a certain threshold, the confidence interval becomes sufficiently small, ensuring that the UCB solution is deemed sufficiently accurate.
\hfill$\Box$ ~\looseness=-1

Algorithm \ref{alg: multiple action} indicates that the algorithm no longer seeks human recourse when $\UCB_t(x^U,a^U)-\LCB_t(x^U,a^U) \leq \Delta$. In other words, for any context $x^U = (x_I^U, x_M^U)$ and action $a$, if the confidence interval falls below $\Delta$, the algorithm stops requesting human input. We show that as more data is collected, the confidence interval shrinks. We want to emphasize that the conclusion holds when the smallest eigenvalue of the covariate matrix of each action grows at a rate of $\beta > 0$, where a larger $\beta$ leads to fewer instances of human intervention. This condition can be met in various scenarios. For instance, if the probability of selecting each action remains strictly positive, the condition holds with $\beta=1$. Essentially, this ensures that information is gathered efficiently, allowing the estimator to converge at a suitable rate for accurate estimation. ~\looseness=-1

Next, by summing over one-step recourse regret derived in Theorem \ref{thm: warm start multiple actions}, we show the cumulative recourse regret. ~\looseness=-1

\begin{theorem}[Improvement Guarantee]\label{thm: improvementguarantee}
    Under the same condition as Theorem \ref{thm: warm start multiple actions}, with probability at least $1-\delta$, the cumulative recourse regret can be bounded by
    \[
\begin{aligned}
\regret(T) 
&\leq 
2 \min \Big\{ \eta T, (\zeta+1) \sqrt{ 2d KT \log\left(\frac{d+T\beta_\mathcal{X}^2}{d}\right) } \\
& \Big( \beta_{\Theta}+\sqrt{2\log\left(\frac{K}{\delta}\right)+d\log\left(1+\frac{T \beta_\mathcal{X}}{d}\right)} \Big) \Big\}.
\end{aligned}
\]
\end{theorem}

Theorem \ref{thm: improvementguarantee} shows that the recourse regret of \textsf{HR-Bandit} is of order $\tilde{O}(\min\{\eta T, d\sqrt{KT}\})$. We analyze two limiting scenarios: When $\eta$ is very large: This corresponds to a low level of human expertise, where human input provides minimal value. In this case, the regret is of order $d\sqrt{KT log K}$, which matches the regret lower bound for the contextual bandits problem. When $\eta=0$: This represents a scenario where the human possesses perfect knowledge. In both cases, the regret achieved by our algorithm is consistent with the theoretical lower bound.  Compared to Theorem \ref{thm: regret}, \textsf{HR-Bandit} offers significant improvement over pure-AI actions, particularly when the problem dimension or action set is large, human action quality is high, and the time horizon is relatively short. However, as the time horizon increases, pure-AI actions will eventually outperform human actions because the model becomes more accurate as more data is gathered.

Note that a larger $\eta$ implies a low human expertise, and the regret will be dominated by the pure AI’s regret. The warm-start guarantee only exists when $\eta$ is small. Consider a scenario where the expert's selections are random or even adversarial, leading to arbitrarily poor decisions. In such cases, the optimal strategy is to rely solely on the AI for decision-making, as the expert's input effectively introduces random noise. Consequently, when the expert's expertise level is low, no algorithm can provide a meaningful warm-start guarantee.

Although Theorem \ref{thm: improvementguarantee} provides strong guarantees on improvement, it relies on certain regularity conditions. However, even when these conditions do not hold for the confidence intervals, we can still ensure that the regret performance of \textsf{HR-Bandit} is no worse than that of the standalone AI algorithm, even if the human collaborator exhibits poor decision-making. This guarantee underlines the robustness of the system, safeguarding its performance against suboptimal human inputs. 
We formally show it in \Cref{thm: robustguarantee}. 
~\looseness=-1

\begin{theorem}[Robustness Guarantee]\label{thm: robustguarantee}
With probability at least $1-\delta$, the cumulative recourse regret of \textsf{HR-Bandit} is bounded that
\[
\begin{aligned}
\regret(T) =\tilde{O}(d\sqrt{KT}).
\end{aligned}
\]
\end{theorem}

\section{Numerical Experiments}

In this section, we demonstrate the efficacy of the proposed methods on both synthetic and real-world datasets. We denote our algorithms in \Cref{alg:cap} and \Cref{alg: multiple action} as \textsf{RLinUCB} and \textsf{HR-Bandit}, respectively. We compare the proposed methods with Linear Upper Confidence Bound (LinUCB) \citep{chu2011contextual}. We use the $\ell_2$-norm as the distance metric in all experiments. 
The implementation is available at \url{https://github.com/ruijiang81/HR-Bandit}. 
~\looseness=-1

\noindent \textbf{Human Behavior Model:} We assume the human experts' decision quality is characterized by the parameter $q$. For each arriving decision instance, the human decision maker will select the optimal recourse and action with probability $q$, and output a random recourse and action with probability $1-q$. We set $q=0.9$, $\Delta=1$, $\gamma=1$, and $\zeta=3$ in the experiments. The results are averaged over 5 runs. We examine the impact of hyperparameters in \Cref{sec:ablation}.

\subsection{Synthetic Data}

We sample $x \sim \mathcal{N}(0,I_d)$, where $d = 5$ and $I_d$ is a $d$-dimensional identity matrix to simulate patient contexts. Here we make the assumption that all features are mutable. The reward $r(a,x) = \theta_a^T x + \epsilon$, where $a\in\{1,2\}$ and $\epsilon \sim \mathcal{N}(0,1)$. The model parameter $\theta_a \sim \mathcal{N}(0,I_d)$. For each time step, we need to decide what recourse and which action to take for the arriving patient with feature $x_t$. Following most algorithmic recourse papers \citep{karimi2020survey,gao2023impact}, we assume that patients will adhere to the recourse recommendations that satisfy the distance constraint.  

\begin{figure}[h]
  \centering
  \begin{subfigure}[b]{0.23\textwidth}
    \includegraphics[width=\textwidth]{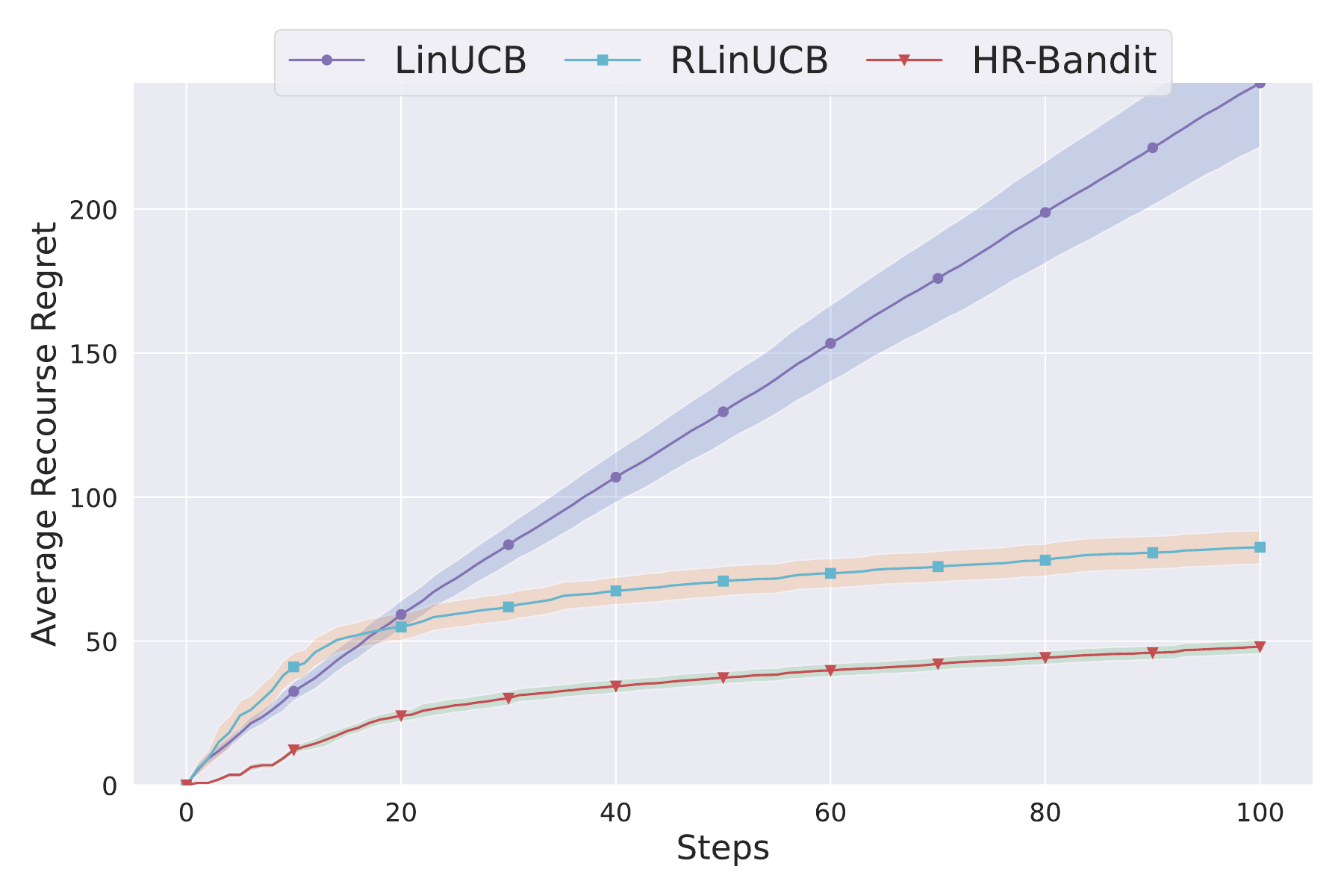}
    \caption{Recourse Regret}
    \label{fig:syn_regret}
  \end{subfigure}
  \hfill 
  \begin{subfigure}[b]{0.23\textwidth}
    \includegraphics[width=\textwidth]{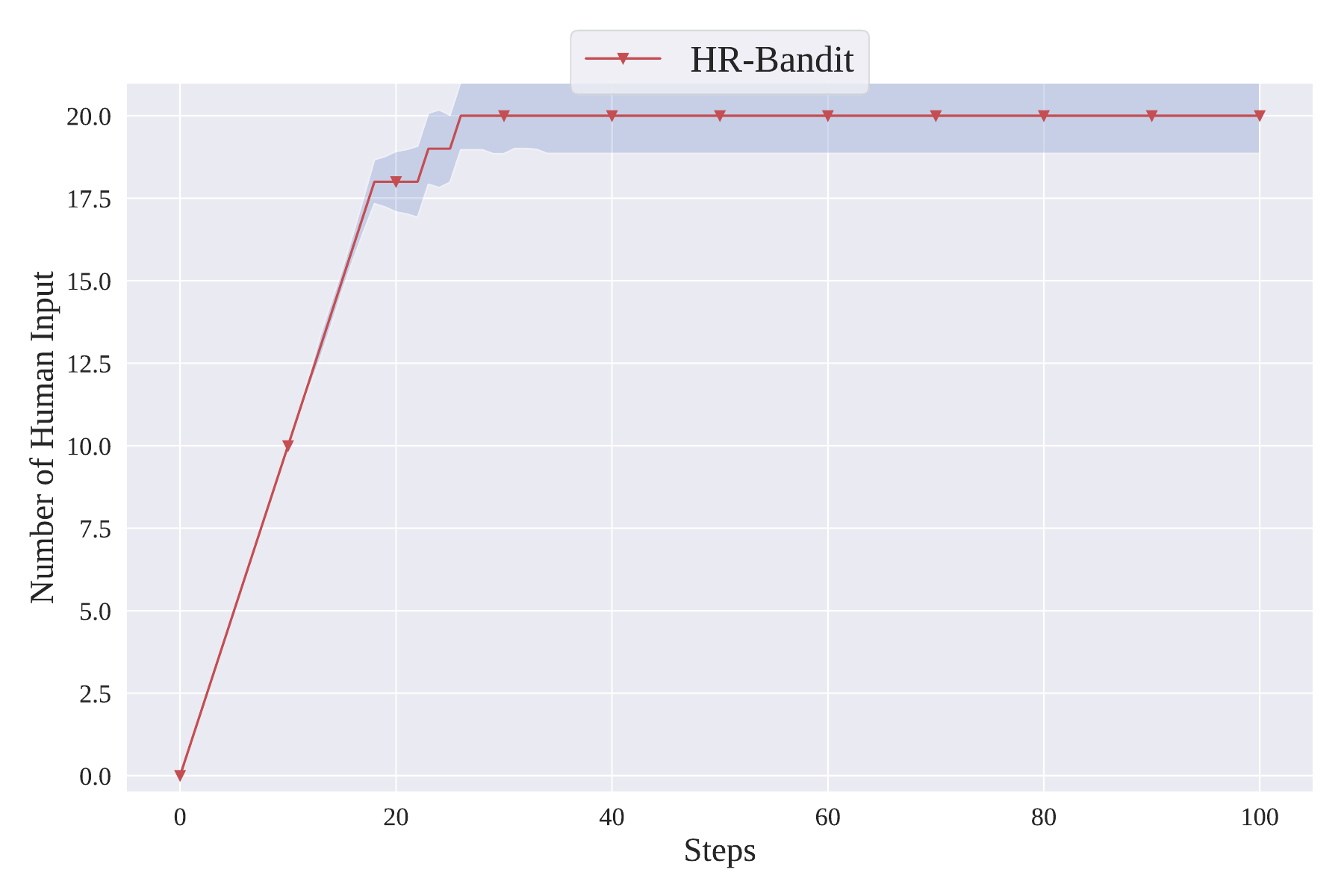}
    \caption{Human Queries}
    \label{fig:syn_human}
  \end{subfigure}
  \caption{Synthetic Data.}
  \label{fig:syn}
\end{figure}

The results of the recourse regret and the number of queries of humans are shown in \Cref{fig:syn_regret} and \Cref{fig:syn_human} respectively. Since LinUCB cannot select recourses, it has a linear recourse regret. \textsf{RLinUCB} significantly improves LinUCB and has a sublinear regret as indicated by \Cref{thm: regret}, which suggests the model learns to prescribe good recourses. \textsf{HR-Bandit} achieves the best regret compared to other methods by utilizing queries from human experts, which confirms \Cref{thm: improvementguarantee}. \Cref{fig:syn_human} shows the system only requires limited number of human inputs, which verifies \Cref{thm: human effort} empirically. 

\subsection{Semi-Synthetic Data}

We build a  healthcare example using real-world data on seminal quality in this section. The real-world dataset is an observational data where part of the patients has received surgical intervention and other patients have not. For our experiments, we need the potential outcomes of one patient for both under surgical treatment or not, also known as counterfactuals. Such counterfactuals are always not available for observational dataset since we can only observe the outcome under one possible treatment in the past. Therefore, we simulate the counterfactuals by fitting a linear model on each treatment arm. 

\citet{gil2012predicting} ask 100 volunteers to provide semen samples and analyze them according to the World Health Organization (WHO) 2010 criteria.  The dataset is publicly available\footnote{https://archive.ics.uci.edu/dataset/244/fertility}. In the dataset, the authors record the age, childish diseases, accident, surgical intervention, high fevers in the last year, alcohol consumption, smoking habit, number of hours sitting per day, and the diagnosis. The outcome variable is any alteration in the sperm parameters, which include Normozoospermia, Asthenozoospermia, Oligozoospermia, and Teratozoospermia.

We consider three mutable features: alcohol consumption, smoking habit, and the number of hours sitting per day. We assume the outcome function is $\mathcal{N}(5,1) * \mathds{1}[\text{diagnosis} = \text{Normal}] + \mathcal{N}(0,1) * \mathds{1}[\text{diagnosis} = \text{Altered}].$ The treatment is whether to ask patients to perform a surgical intervention. 
To simulate the counterfactual outcome when patients implement recourse recommendations that are not in the dataset, we fit separate linear regression models for each treatment arm and use them as the underlying data-generating processes for the corresponding arm. 

\begin{figure}[h]
  \centering
  \begin{subfigure}[b]{0.23\textwidth}
    \includegraphics[width=\textwidth]{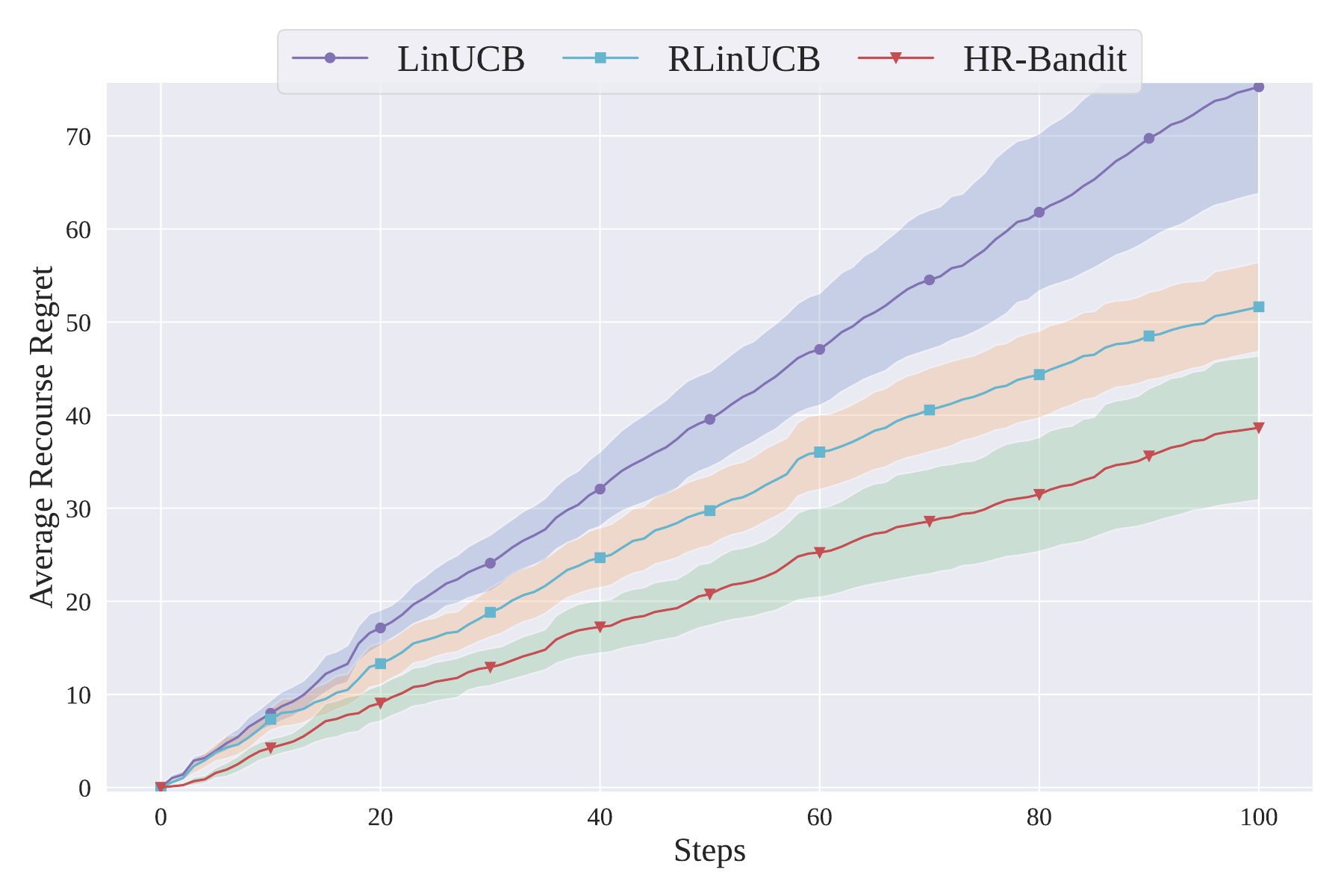}
    \caption{Recourse Regret}
    \label{fig:fert_regret}
  \end{subfigure}
  \hfill 
  \begin{subfigure}[b]{0.23\textwidth}
    \includegraphics[width=\textwidth]{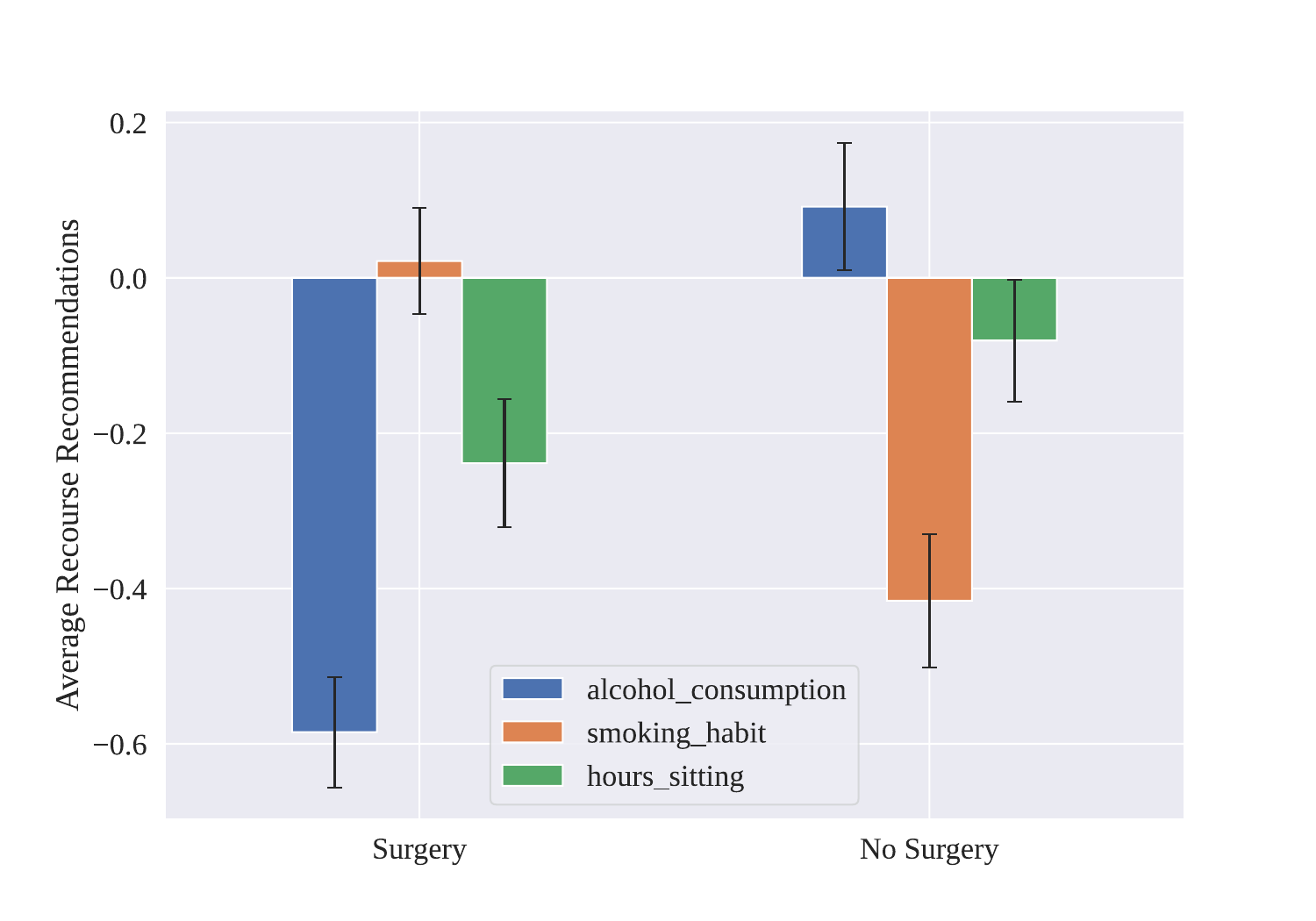}
    \caption{Recourses Selected}
    \label{fig:recourses}
  \end{subfigure}
  \caption{Fertility Data.}
  \label{fig:fert}
\end{figure}

The recourse regret is shown in \Cref{fig:fert_regret}. The number of humans queried is included in \Cref{app:addexp}. We observe similar qualitative conclusions as the synthetic data where \textsf{HR-Bandit} significantly outperforms other methods with only limited number of human queries. In addition, we visualize the recourses output after step 70 by \textsf{HR-Bandit} in \Cref{fig:recourses} (We note that while we did not restrict recourses to be negative, our algorithm can easily accommodate this constraint if needed).  We choose step 70 to show the recourse recommendation at a later stage since in the beginning there will be more explorations than exploitations.
The coefficients of alcohol consumption, smoking habit, and the number of hours sitting per day fitted by the linear regression are $[-0.31, -0.03, -0.15]$ and $[-0.03, -0.17, -0.12]$ for the surgery and no surgery groups, respectively. As a result, we observe that our algorithm encourages the patients to focus more on reducing alcohol consumption and sitting hours if the algorithm is going to recommend surgery and more on smoking if the algorithm decides surgery is not necessary. ~\looseness=-1

In addition, we perform experiment on the Infant Health and Development Program (IHDP) dataset \citep{hill2011bayesian} in \Cref{app:addexp} with similar qualitative conclusions. 

\subsection{Ablation Studies}
\label{sec:ablation}

\begin{figure}[h]
  \centering
  \begin{subfigure}[b]{0.23\textwidth}
    \includegraphics[width=\textwidth]{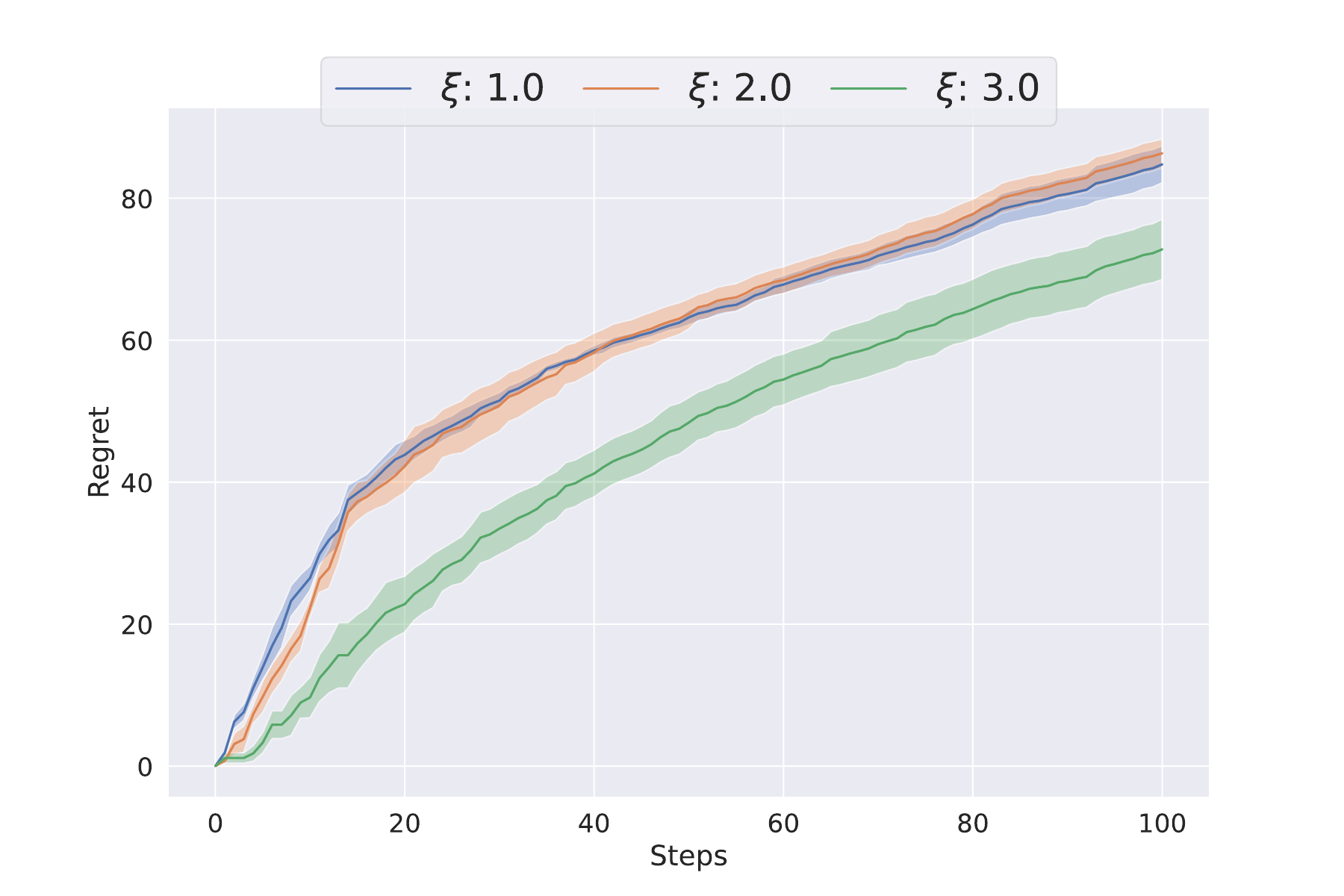}
    \caption{Recourse Regret}
    \label{fig:humantrust_regret}
  \end{subfigure}
  \hfill 
  \begin{subfigure}[b]{0.23\textwidth}
    \includegraphics[width=\textwidth]{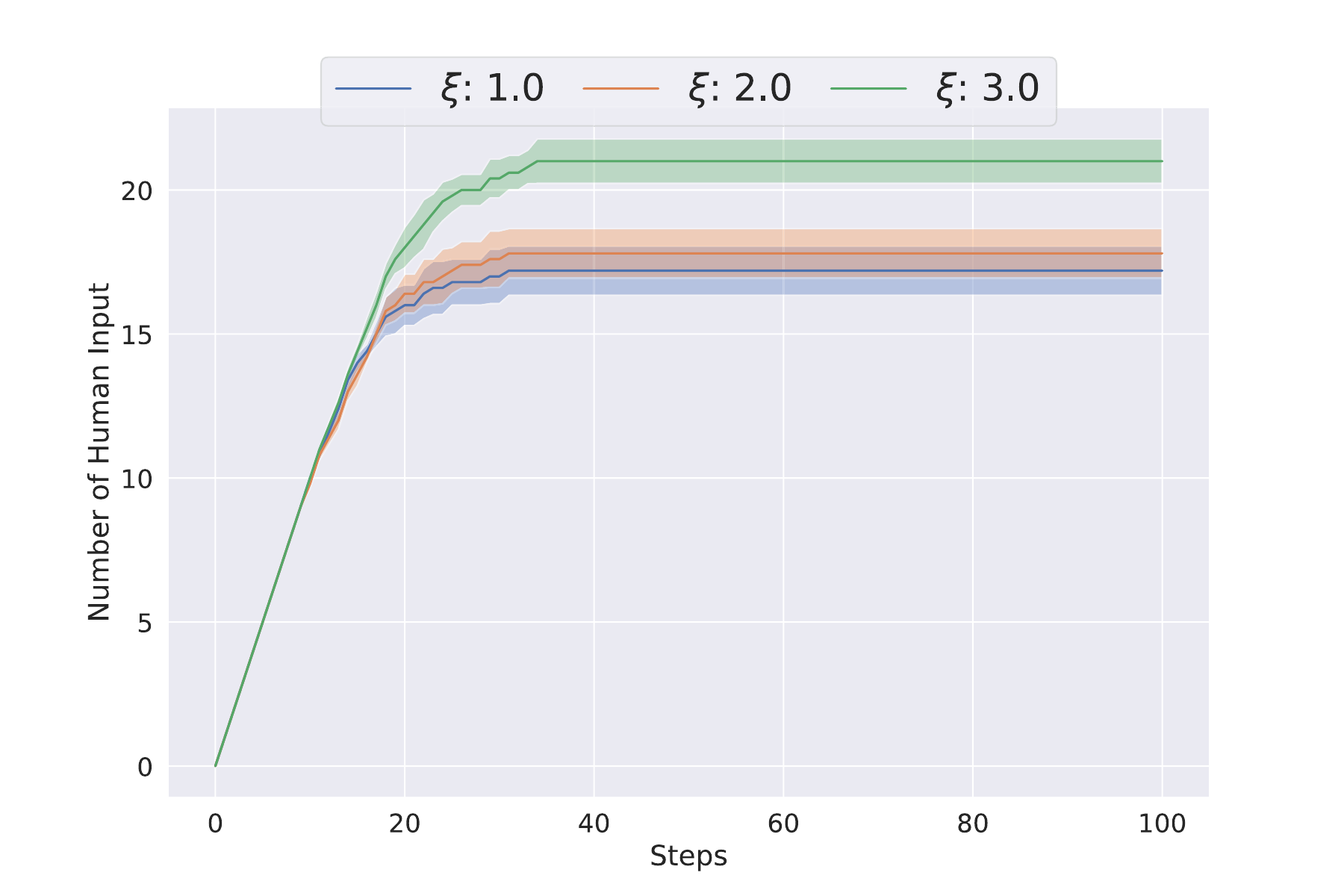}
    \caption{Human Queries}
    \label{fig:humantrust_human}
  \end{subfigure}
  \caption{Impact of $\zeta$ (Human Trust).}
  \label{fig:ablation_trust}
\end{figure}

\begin{figure}[h]
  \centering
  \begin{subfigure}[b]{0.23\textwidth}
    \includegraphics[width=\textwidth]{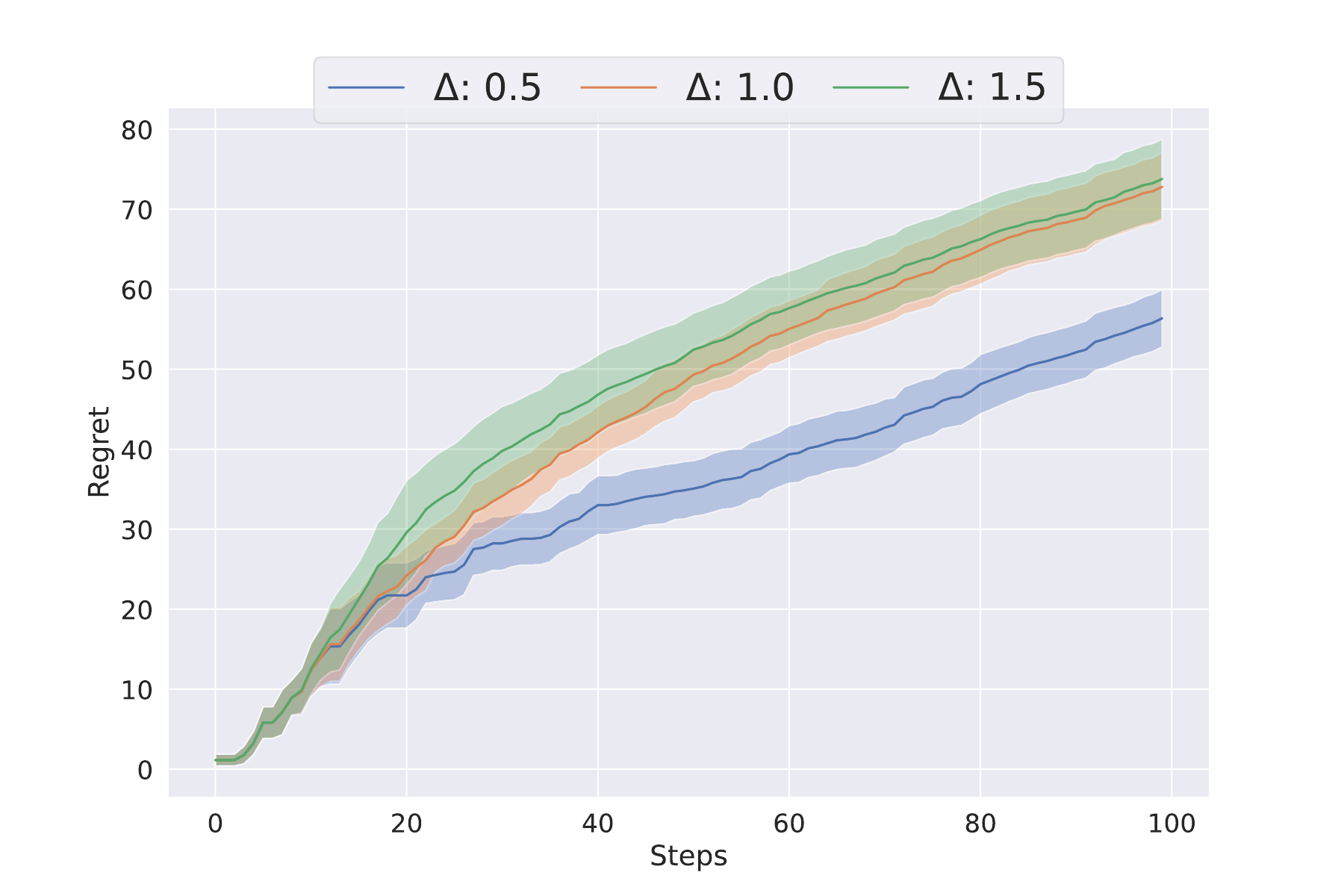}
    \caption{Recourse Regret}
    \label{fig:humanask_regret}
  \end{subfigure}
  \hfill 
  \begin{subfigure}[b]{0.23\textwidth}
    \includegraphics[width=\textwidth]{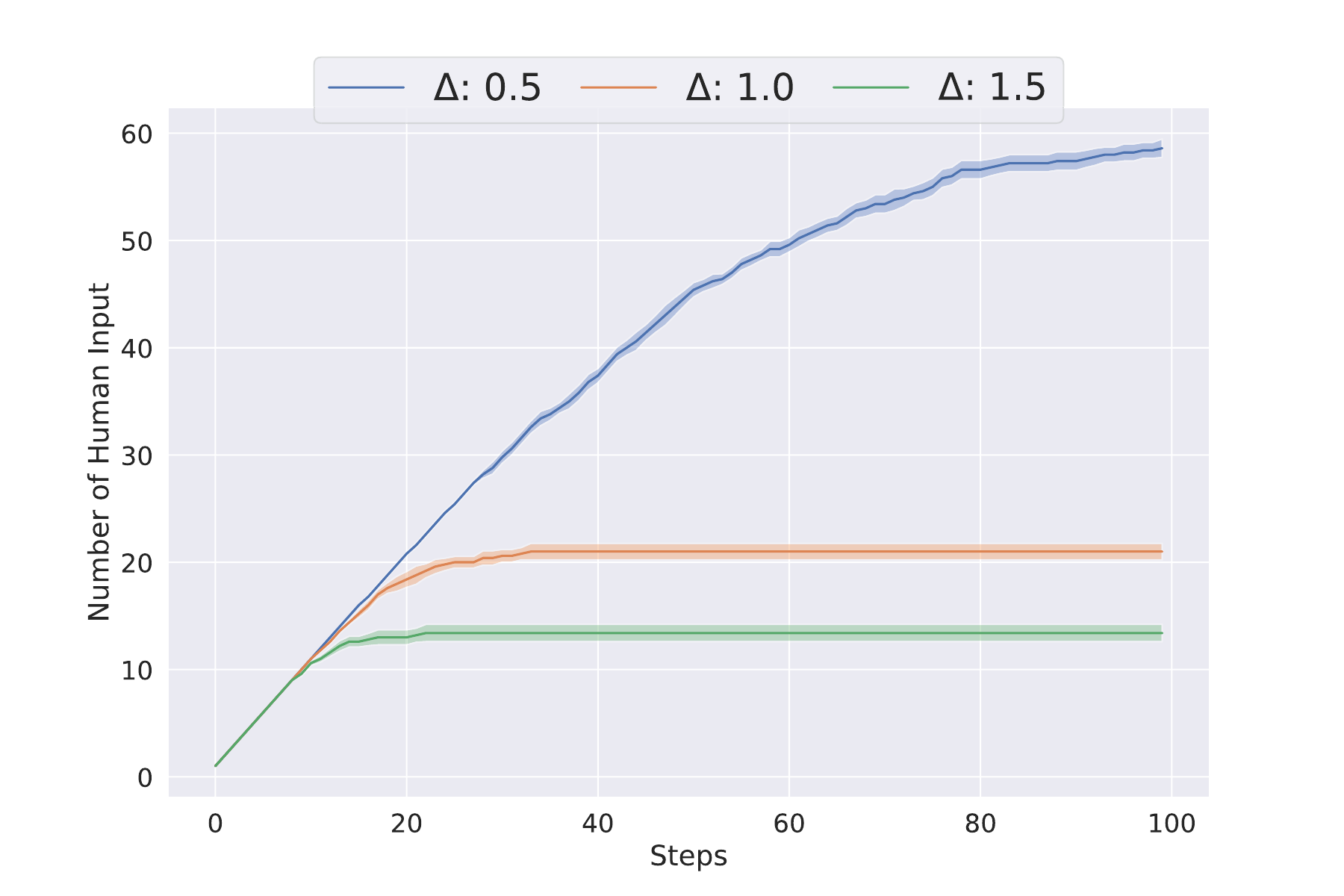}
    \caption{Human Queries}
    \label{fig:humanask_human}
  \end{subfigure}
  \caption{Impact of $\Delta$ (Human Effort).}
  \label{fig:ablation_ask}
\end{figure}

\begin{figure}[!htbp]
  \centering
  \begin{subfigure}[b]{0.23\textwidth}
    \includegraphics[width=\textwidth]{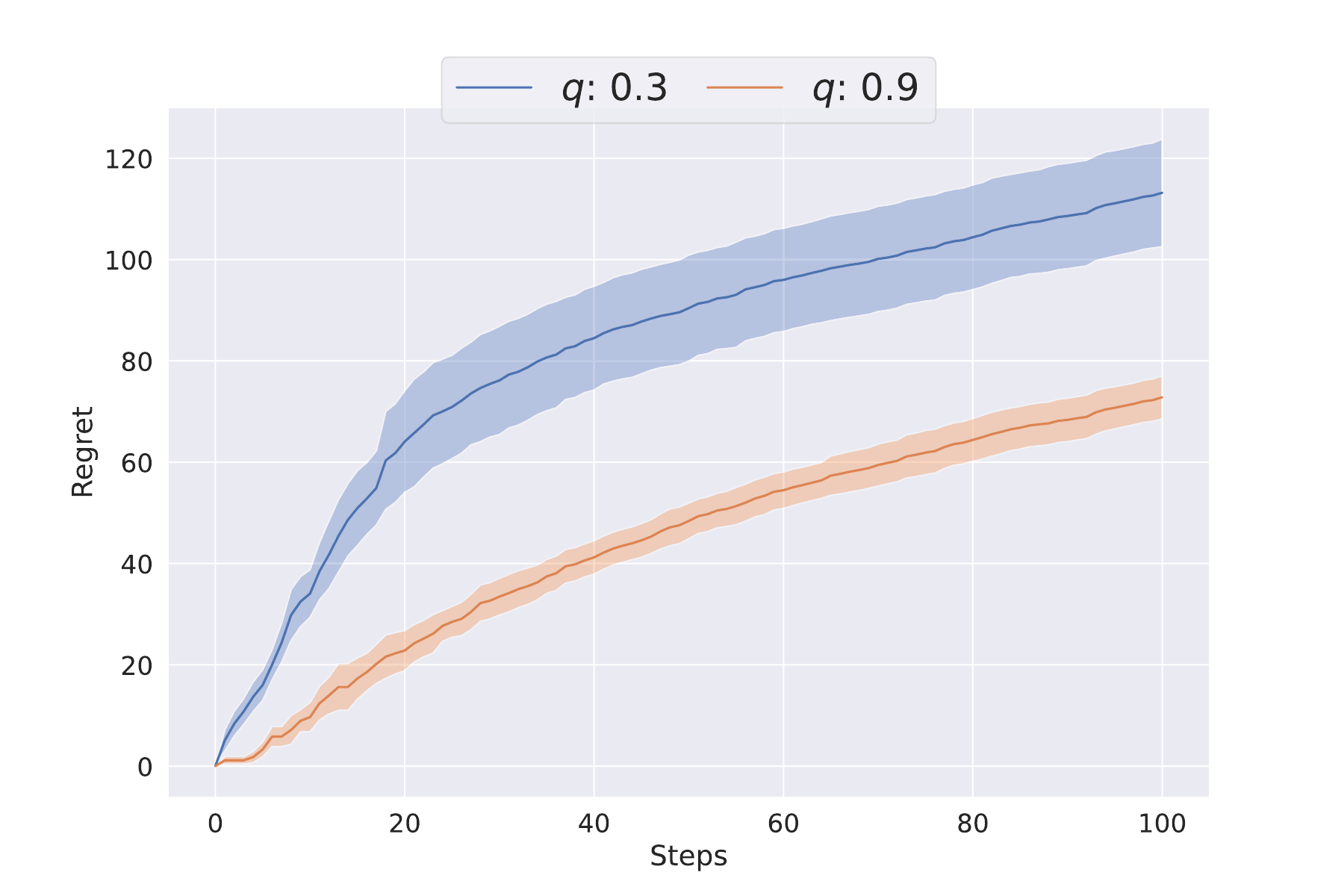}
    \caption{Recourse Regret}
    \label{fig:humanq_regret}
  \end{subfigure}
  \hfill 
  \begin{subfigure}[b]{0.23\textwidth}
    \includegraphics[width=\textwidth]{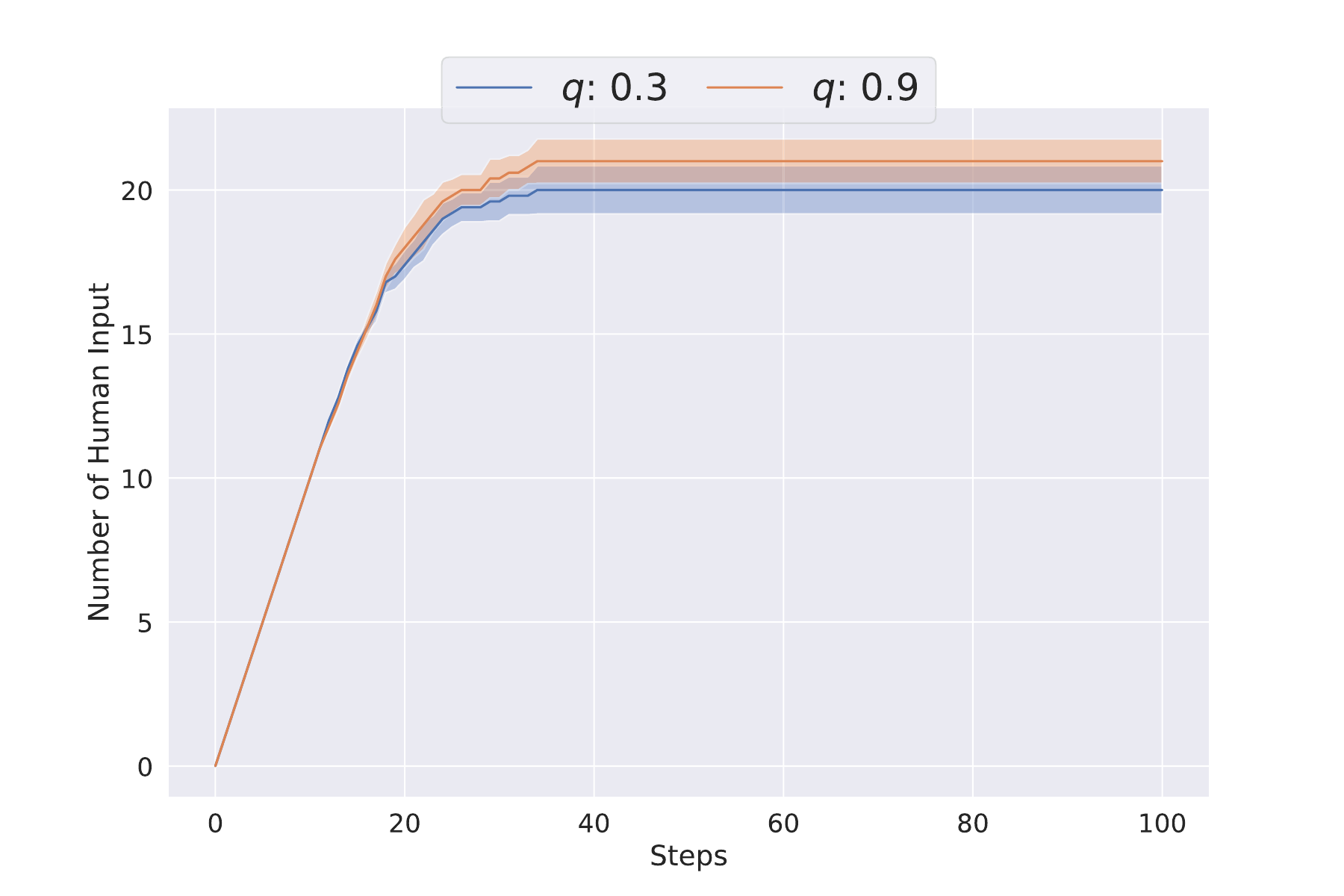}
    \caption{Human Queries}
    \label{fig:humanq_human}
  \end{subfigure}
  \caption{Impact of $q$ (Human Expertise).}
  \label{fig:ablation_q}
\end{figure}

Here we examine the impact of $\zeta$, $\Delta$ in \Cref{alg: multiple action}, and the human quality $q$ in \Cref{fig:ablation_trust}, \Cref{fig:ablation_ask}, and \Cref{fig:ablation_q} respectively. $\zeta$ is the variance control parameter that indicates our trust in the human experts and $\Delta$ controls how often we want to consult the human experts. 

As $\zeta$ increases, we observe that \textsf{HR-Bandit} gains more benefit by leveraging additional input from the human expert. Conversely, when $\Delta$ is larger, the system stops consulting the human earlier, potentially resulting in higher regret if the human expert is of high quality. Finally, we evaluate the robustness of \textsf{HR-Bandit} under poor human decision quality by setting $q=0.3$. Consistent with \Cref{thm: improvementguarantee}, \textsf{HR-Bandit} maintains sub-linear regret even with low-quality human inputs, highlighting the safety and reliability of our system. 

\section{Conclusion and Future Work}

In this work, we introduced \textsf{RLinUCB} and \textsf{HR-Bandit}, extending algorithmic recourse and human-AI collaboration to online learning with theoretical guarantees. \textsf{HR-Bandit} offers three key properties: (i) improvement guarantee, (ii) human-effort guarantee, and (iii) robustness guarantee. Empirical results show its superior performance over standard bandit algorithms. However, our approach is not without limitations. 
First, we only consider linear reward functions and future work could explore non-linear reward structures. Second, while we follow the common assumption in the recourse literature that patients will follow recourse recommendations \citep{gao2023impact,verma2020counterfactual}, in practice, patients may not follow the suggestions entirely. Future work can model patient noncompliance behaviors to better capture real-world complexities.

\newpage
\bibliography{aistats25/ref}
\newpage 

\onecolumn 
\appendix 
\section{Proofs}\label{appendix: proof}
\label{app:proof}

\begin{proof}[Proof of Lemma~\ref{lemma: optimal solution}]
We first write down the Lagrangian relaxation for \eqref{E. optimization} for $\lambda > 0$ as follows:
\[
\mathcal{L}(\checkx_M; \lambda) =\checkx_M^\top \theta_{a,M}^* + \lambda (\gamma^2 - \|\checkx_M- x_M\|_2^2).
\]
Then, a Karush–Kuhn–Tucker (KKT)  point for the optimization model \eqref{E. optimization} is some point $(\checkx_M^*, \lambda^*)$ such that it satisfies the following conditions:
\[
\begin{aligned}
\nabla_{\checkx_M^*} \mathcal{L}& (\checkx_M^*; \lambda)  = \; \theta_{a,M}^* - 2 \lambda^* (\|\checkx_M^* - x_M\|_2) = 0,\\
&\lambda^* \cdot (\gamma^2 - \|\checkx_M^* - x_M\|_2^2) = 0, \\
& \|\checkx_M^* - x_M\|_2^2 \leq \gamma^2.
\end{aligned}
\]
Then, it can be observed that the solution $\checkx_M^*=x_M+\theta_{a,M}^*/\|\theta_{a,M}^*\|\gamma$ and $\lambda^* = \theta/2\gamma$ satisfies the above KKT conditions, which implies that it is an optimal solution to the counterfactual optimization problem \eqref{E. optimization}.
\end{proof}

The proof of Lemma \ref{lemma: optimal solution2} follows similarly.

The following lemma is based on Theorem 20.5 in \citep{lattimore2020bandit}.
\begin{lemma}\label{lemma: radius}
    Let $\delta\in (0,1)$. Then, with probability at least $1-\delta$, for all $t\in [T]$ and all $a\in \mathcal{A}$, it holds that 
    \[
    \|\hat{\theta}_{ta}-\theta_a^*\|_{V_{ta}}\leq \beta_{\Theta}+\sqrt{2\log\left(\frac{1}{\delta}\right)+\log\left(\det (V_{ta})\right)}
    \]
\end{lemma}
By the union bound, we have $ \|\hat{\theta}_{ta}-\theta_a^*\|_{V_{ta}}\leq \rho_{ta}$ holds for all $a\in \mathcal{A}$ with probability at least $1-\delta$, where 
\[
\rho_{ta}=\beta_{\Theta}+\sqrt{2\log\left(\frac{K}{\delta}\right)+d\log\left(1+\frac{\sum_{s=1}^{t-1} \mathds{1}(a_t=a) \beta_\mathcal{X}}{d}\right)}.
\]

\begin{proof}[Proof of Theorem \ref{thm: regret}.]
Assume the optimal action selected at time $t$ is $a_t^*$. Recall that the regret is defined as
\[
\regret_\pi(T) = E_\pi\left[\sum_{t=1}^T r(x_t^*, a_t^*) - r(x_t, a_t) \right].
\]
Since $x_t$ and $\theta_{ta}$ is the optimal solution to \eqref{E. optimization2}, we have
\[
\theta_{ta}^\top x_t\geq \theta_{a_t^*}^\top x_t^* \geq \theta_{a_t^*}^\top x_t\geq \theta_{ta^*_t}^\top x_t.
\]
Then the one-step regret can be bounded by 
\[
r(x_t^*, a_t^*) - r(x_t, a_t) \leq \theta_{ta^*_t}^\top x_t- \theta_{ta}^\top x_t\leq\rho_{ta} \|x_t\|_{V_{ta}^{-1}}.
\]
Therefore, 
\[
\begin{aligned}
\regret_\pi(T)&\leq \sum_{t=1}^T \left(\sum_{a\in \mathcal{A}}\rho_{ta} \|x_t\|_{V_{ta}^{-1}} 1(a_t=a)\right)\\
&\leq  \rho_T\sum_{t=1}^T \sum_{a\in \mathcal{A}} \|x_t\|_{V_{ta}^{-1}} 1(a_t=a).
\end{aligned}
\]
Applying the elliptical potential lemma \cite[Lemma 19.4]{lattimore2020bandit}, for each $a\in \mathcal{A}$, we have that
\[
\begin{aligned}
&\sum_{t=1}^{T} \sum_{a\in \mathcal{A}} \| x_t 1(a_t=a)\|_{V_{ta}^{-1}}^2\\
\leq &\sum_{t=1}^{T}\sum_{a\in \mathcal{A}}  1\wedge \|x_t 1(a_t=a)\|_{V_{ta}^{-1}}^2\\
\leq &2 dK\log \left(\frac{t }{d \lambda_{\min}^{1/d}}\right)\leq 2 d K\log\left(\frac{d+T\beta_\mathcal{X}^2}{d}\right)\,.
\end{aligned}
\]
Thus, we reach our conclusion that
\[
\begin{aligned}
\regret_\pi(T)&\leq  \rho_T \sqrt{2 d KT\log\left(\frac{d+T\beta_\mathcal{X}^2}{d}\right)}.
\end{aligned}
\]
\end{proof}

\begin{proof}[Proof of Theorem \ref{thm: warm start multiple actions}.]
  There are two scenarios in the implementation of Algorithm \ref{thm: warm start multiple actions}: Whether it holds that $\UCB(x^U,a^U)-\LCB(x^U,a^U)>\Delta$ where $\Delta$ is taken the value of $\eta$. 

1. When $\UCB(x^U,a^U)-\LCB(x^U,a^U)>\Delta=\eta$, then we ask human for recourse $x^H$, and it satisfies that
 $r(x^H,a^H)\geq r(x^*,a^*)-\eta$ according to the condition. 
 
 \underline{When human's action is taken}, note that
 \[
 \begin{aligned}
 \UCB(x^U,a^U)\geq \UCB(x^*,a^*)\geq r(x^*,a^*)\geq r(x^U,a^U)\geq \LCB(x^U,a^U)
 \end{aligned}
 \]
 and
 \[
 \UCB(x^U,a^U)\leq  \UCB(x^H,a^H)+ 2\CI(x^U,a^U).
 \]
 It implies that
 \[
  \begin{aligned}
 &\regret(x^H,a^H)=r(x^*,a^*)-r(x^H,a^H)\\
 &\leq  \UCB(x^U,a^U)-\LCB(x^H,a^H)\\
 &\leq \UCB(x^H,a^H)+2\CI(x^U,a^U)-\UCB(x^H,a^H)+2\CI(x^H,a^H)\\
 &\leq 2(1+\zeta)\min\{\CI(x^U,a^U), \CI(x^H,a^H)\}.
  \end{aligned}
 \]
 Thus, we have 
 \[
 \regret(x^H,a^H)\leq \min\{\eta, 2(1+\zeta)\CI(x^H,a^H)\},
 \]
 and also we have
 \[
  \regret(x^H,a^H)\leq \min\{\eta, 2(1+\zeta)\CI(x^U,a^U)\}.
 \]
 \underline{When human's action is not taken}, then 
  \[
  \begin{aligned}
 &\regret(x^U,a^U)\leq  2\CI(x^U,a^U).
  \end{aligned}
 \]
Under the condition stated in the theorem,  it  holds that $\UCB(x^H,a^H)\leq \LCB(x^U,a^U)$, so it also satisfies that 
 \[
 \begin{aligned}
 r(x^U,a^U)\geq \LCB(x^U,a^U)\geq\UCB(x^H,a^H)\geq  r(x^H,a^H)\geq r(x^*,a^*)-\eta.
 \end{aligned}
 \]
 Thus, we have 
  \[
 \regret(x^U,a^U)\leq \min\{\eta, \CI(x^U,a^U)\}.
 \]

2. When $\UCB(x^U,a^U)-\LCB(x^U,a^U)\leq \Delta$, then human does not propose recourse, and we have 
 \[
  \begin{aligned}
 \regret(x^U,a^U)\leq \UCB(x^U,a^U)-\LCB(x^U,a^U)\leq 2\CI(x^U,a^U)\leq 2\eta.
  \end{aligned}
 \]
In this case, it also satisfies that
\[
\regret(x^U,a^U)\leq 2\min\{\eta, \CI(x^U,a^U)\}.
\]
To sum up, 
 \[
\regret(x^U,a^U)\leq 2\min\{\eta, (1+\zeta)\CI(x^U,a^U)\}.
\]
\end{proof}

\begin{proof}[Proof of Theorem \ref{thm: human effort}.]
Algorithm \ref{alg: multiple action} suggests that, human does not propose recourse anymore if $\UCB_t(x^U,a^U)-\LCB_t(x^U,a^U)\leq \Delta$.  That is, if for any context $x^U=(x_I^U, x_M^U)$ and action $a^U$, 
\[
\UCB_t(x^U, a^U)-\LCB_t(x^U,a^U)=2\rho_{t a}\|x^U\|_{V_{ta}^{-1}}\leq \Delta,
\]

which is equivalent to that 
\[
\beta_{\Theta}+\sqrt{2\log\left(\frac{K}{\delta}\right)+d\log\left(1+\frac{\sum_{s=1}^{t-1} \mathds{1}(a_t=a) \beta_\mathcal{X}}{d}\right)}\leq \frac{\Delta}{2\|x^U\|_{V_{ta}^{-1}}},
\]
then human's proposal of action $x^H$ will never be selected.

Let $n_{ta}=\sum_{s=1}^{t-1} \mathds{1}(a_t=a)$. According to the condition that $\lambda_{\min} (V_{ta})\geq \omega n_{ta}^\beta$, we have
\[
\|x^U\|_{V_{ta}^{-1}}^2\leq \frac{\beta_{\mathcal{X}}^2}{\lambda_{\min}(V_{ta})}\leq \frac{\beta_{\mathcal{X}}^2}{\omega n_{ta}^{\beta}}.
\]
Therefore, as long as
\[
\beta_{\Theta}+\sqrt{2\log\left(\frac{K}{\delta}\right)+d\log\left(1+\frac{n_{ta} \beta_\mathcal{X}}{d}\right)}\leq \frac{\Delta \sqrt{\omega} n_{ta}^{\beta/2}}{2 \beta_{\mathcal{X}}},
\]
we do not need human's advice anymore.

When 
\[
n_{ta}\geq \left(\frac{6\beta_{\Theta}\beta_{\mathcal{X}}}{\Delta \sqrt{\omega}}\right)^{\frac{2}{\beta}},
\]
we can bound the first term $\beta_{\Theta}$ by the 1/3 of the right-hand-side term, i.e., 
\[
\beta_\Theta\leq \frac{1}{3}\cdot \frac{\Delta \sqrt{\omega} n_{ta}^{\beta/2}}{2 \beta_{\mathcal{X}}}.
\]
When
\[
n_{ta}\geq \left(\frac{12\log(K/\delta)\beta_{\mathcal{X}}}{\Delta\sqrt{\omega}}\right)^{\frac{2}{\beta}},
\]
we can bound the second term by 
\[
2\log(K/\delta)\leq \frac{1}{3}\cdot \frac{\Delta \sqrt{\omega} n_{ta}^{\beta/2}}{2 \beta_{\mathcal{X}}}.
\]
When 
\[
\log(1+\frac{\beta_{\mathcal{X}}}{d} n_{ta})\leq \frac{1}{3}\cdot \frac{\Delta \sqrt{\omega} n_{ta}^{\beta/2}}{2d \beta_{\mathcal{X}}},
\]
we can bound the third term by 
\[
d\log\left(1+\frac{n_{ta} \beta_\mathcal{X}}{d}\right)\leq  \frac{1}{3}\cdot \frac{\Delta \sqrt{\omega} n_{ta}^{\beta/2}}{2 \beta_{\mathcal{X}}}.
\]
Define $\bar{n}=\max\{n': \log(1+\frac{\beta_{\mathcal{X}}}{d} n')\leq  \frac{\Delta \sqrt{\omega} n'{}^{\beta/2}}{6d \beta_{\mathcal{X}}}\}$. Then, when 
\[
n_{ta}\geq \max\left\{\left(\frac{6\beta_{\Theta}\beta_{\mathcal{X}}}{\Delta \sqrt{\omega}}\right)^{\frac{2}{\beta}}, \left(\frac{12\log(K/\delta)\beta_{\mathcal{X}}}{\Delta\sqrt{\omega}}\right)^{\frac{2}{\beta}}, \bar{n}\right\}=: n^{\diamond},
\]
we  can conclude that
\[
\begin{aligned}
&\beta_{\Theta}+\sqrt{2\log\left(\frac{K}{\delta}\right)+d\log\left(1+\frac{n_{ta} \beta_\mathcal{X}}{d}\right)}\\
\leq &\beta_{\Theta}+2\log\frac{K}{\delta}+d\log\left(1+\frac{n_{ta} \beta_\mathcal{X}}{d}\right)\\
\leq &\frac{\Delta \sqrt{\omega} n_{ta}^{\beta/2}}{2 \beta_{\mathcal{X}}}.
\end{aligned}
\]

Therefore, summing over all actions, the maximum time that human's recourse will be adopted is $Kn^{\diamond}$.
\end{proof}

\begin{proof}[Proof of Theorem \ref{thm: improvementguarantee}.] We first define the regret of the proposed algorithm as follows:
\[
\begin{aligned}
\regret(T) = \sum_{t \in [T]} \left( r(x_t^*, a_t^*) - r(x_{t}, a_t) \right),
\end{aligned}
\]
where $x_{t}^*=(x_{tI}^*, \checkx_{tM}^*)$ is the optimal recourse and $a_t^*$ is the optimal action, which are the optimal solution to \eqref{E. optimization} at time step $t$. Also, $(x_{t}, a_t)$ is the recourse and action pair implemented by \textsf{HR-Bandit} at time step $t$.

In the proof of Theorem \ref{thm: warm start multiple actions}, it is proved that when human's action is taken,
 \[
 \regret(x^H,a^H)\leq \min\{\eta, 2(1+\zeta)\CI(x^H,a^H)\}.
 \]
 When human's action is not taken, 
 \[
 \regret(x^U,a^U)\leq  \min\{\eta, 2(1+\zeta)\CI(x^U,a^U)\}.
 \]
Summing over all time steps, we have
\[
\begin{aligned}
\regret(T) &= \sum_{t \in [T]} \left( r(x_t^*, a_t^*) - r(x_{t}, a_t) \right)\\
&\leq \min\{\eta, 2(1+\zeta)\CI_t(x_t,a_t)\}.
\end{aligned}
\]

We can then make the following argument to bound the expression on the right-hand side of the above bound:
\[
\begin{aligned}
\sum_{t \in [T] 
} 
\CI_t(x_t,a_t)
&\leq 
\left( \beta_{\Theta}+\sqrt{2\log\left(\frac{K}{\delta}\right)+d\log\left(1+\frac{T \beta_\mathcal{X}}{d}\right)} \right)
\cdot \Bigg(\sum_{t \in [T]
} \sum_{a \in \mathcal{A}} \| x_{t} \|_{V_{ta}^{-1}} \;\mathds{1}(a_t = a) \Bigg).
\end{aligned}
\]

Using Elliptical potential lemma, we can establish the following argument:
\[
\begin{aligned}
    &\sum_{t \in [T]} \sum_{a \in \mathcal{A}} \|x_{t}\|_{V_{ta}^{-1}} \;\mathds{1}(a_t = a) \\
    &\leq 
    \sqrt{T} \sqrt{\sum_{t \in [T]} \sum_{a \in \mathcal{A}} \|x_{t} \;\mathds{1}(a_t = a) \|^2_{V_{ta}^{-1}} } \\
    &\leq \sqrt{ 2TK \log\left( \frac{\det(V_T)}{\det(V_0)} \right) } \leq \sqrt{ 2dKT  \log\left(\frac{d+T\beta_\mathcal{X}^2}{d}\right) }.
\end{aligned}
\]
Using this bound, we can finally bound the recourse regret as follows:
\[
\begin{aligned}
\regret(T) \leq 
2 \min \Big\{ \eta T, (\zeta+1) \sqrt{ 2d KT \log\left(\frac{d+T\beta_\mathcal{X}^2}{d}\right) } 
\left( \beta_{\Theta}+\sqrt{2\log\left(\frac{K}{\delta}\right)+d\log\left(1+\frac{T \beta_\mathcal{X}}{d}\right)} \right) \Big\},
\end{aligned}
\]
which completes the proof. 

\end{proof}

\begin{proof}[Proof of Theorem \ref{thm: robustguarantee}.]
The proof follows similarly with that in Theorem \ref{thm: warm start multiple actions}.

1. When $\UCB(x^U,a^U)-\LCB(x^U,a^U)>\Delta$, then we ask human for recourse $x^H$.
 
 \underline{When human's action is taken}, note that
 \[
 \begin{aligned}
 &\UCB(x^U,a^U)\geq \UCB(x^*,a^*)\geq r(x^*,a^*)\geq r(x^U,a^U)\geq \LCB(x^U,a^U)
 \end{aligned}
 \]
 and
 \[
 \UCB(x^U,a^U)\leq  \UCB(x^H,a^H)+ 2\CI(x^U,a^U).
 \]
 It implies that
 \[
  \begin{aligned}
 &\regret(x^H,a^H)=r(x^*,a^*)-r(x^H,a^H)\\
 &\leq  \UCB(x^U,a^U)-\LCB(x^H,a^H)\\
 &\leq \UCB(x^H,a^H)+2\CI(x^U,a^U)-\UCB(x^H,a^H)+2\CI(x^H,a^H)\\
 &\leq 2(1+\zeta)\CI(x^H,a^H).
  \end{aligned}
 \]
 Thus, we have 
 \[
 \regret(x^H,a^H)\leq  2(1+\zeta)\CI(x^H,a^H).
 \]
 \underline{When human's action is not taken}, then 
  \[
  \begin{aligned}
 &\regret(x^U,a^U)\leq  2\CI(x^U,a^U).
  \end{aligned}
 \]
2. When $\UCB(x^U,a^U)-\LCB(x^U,a^U)\leq \Delta$, then human does not propose recourse, and we have 
 \[
  \begin{aligned}
 &\regret(x^U,a^U)\leq \UCB(x^U,a^U)-\LCB(x^U,a^U)\leq 2\CI(x^U,a^U).
  \end{aligned}
 \]
In this case, it also satisfies that
\[
\regret(x^U,a^U)\leq 2 \CI(x^U,a^U).
\]
Summing over all time steps, we have
\[
\begin{aligned}
\regret(T) &= \sum_{t \in [T]} \left( r(x_t^*, a_t^*) - r(x_{t}, a_t) \right)\\
&\leq  2(1+\zeta)\CI(x_t,a_t).
\end{aligned}
\]
Following the same steps in the proof for Theorem \ref{thm: improvementguarantee}, we can reach the conclusion that 
\[
\begin{aligned}
\regret(T) \leq 
2  (\zeta+1) \sqrt{ 2d KT \log\left(\frac{d+T\beta_\mathcal{X}^2}{d}\right) } 
\left( \beta_{\Theta}+\sqrt{2\log\left(\frac{K}{\delta}\right)+d\log\left(1+\frac{T \beta_\mathcal{X}}{d}\right)} \right).
\end{aligned}
\]
\end{proof}

\section{Discussion on Assumptions}
\label{app:assumption}
The assumption in Theorem \ref{thm: warm start multiple actions} assumes that there exists a constant $\zeta>0$ such that for any $(x_1, a_1), (x_2, a_2)\in\mathcal{X}$, it holds that $\CI(x_1, a_1)\leq \zeta \CI(x_2,a_2)$. To be more specific, it assumes that $\rho_{ta_1}\|x_1\|_{V_{ta_1}^{-1}}< \zeta \rho_{ta_2}\|x_2\|_{V_{ta_2}^{-1}}$
where
\[
\rho_{ta}=\beta_{\Theta}+\sqrt{2\log\left(\frac{K}{\delta}\right)+d\log\left(1+\frac{ n_{ta} \beta_\mathcal{X}}{d}\right)}.
\]
We claim that when the following two conditions are satisfied, this condition is naturally satisfied:
\begin{enumerate}
    \item There exists a constant $\varrho $ such that $n_{ta_1}/n_{ta_2}\leq \varrho$; \label{condition 1}
    \item There exists a constant $\varrho'$ such that $\lambda_{\min}(V_{t a_1})\geq \varrho' \lambda_{\max}(V_{t a_2})$. \label{condition 2}
\end{enumerate}
First, to see why this claim is true, according to the definition of $\rho_{ta}$, we have
\[
\frac{\rho_{t a_1}}{\rho_{t a_2}}\leq \frac{n_{t a_1}}{n_{t a_2}}\leq \varrho.
\]
For any $x_1$ and $x_2$, it holds that
\[
\|x_1\|_{V_{t a_1}^{-1}}\leq \|x_1\|_2/\sqrt{\lambda_{\min}(V_{ta_1})}\leq \beta_{\mathcal{X}}/\sqrt{\varrho'\lambda_{\max}(V_{ta_2})}\leq\frac{\beta_{\mathcal{X}}}{\underline{\beta}_{\mathcal{X}}\sqrt{\varrho'}} \|x_2\|_2/\sqrt{\lambda_{\max}(V_{ta_2})}\leq \frac{\beta_{\mathcal{X}}}{\underline{\beta}_{\mathcal{X}}\sqrt{\varrho'}}\|x_2\|_{V_{ta_2}^{-1}}.
\]
Thus, we can conclude that 
\[
\rho_{ta_1}\|x_1\|_{V_{ta_1}^{-1}}< \zeta \rho_{ta_2}\|x_2\|_{V_{ta_2}^{-1}},
\]
where 
\[
\zeta =\max\left\{\varrho,\frac{\beta_{\mathcal{X}}}{\underline{\beta}_{\mathcal{X}}\sqrt{\varrho'}}\right\}.
\]
Condition \ref{condition 1} basically ensures that each action is taken a sufficient number of times. %
Condition \ref{condition 2} requires the context to be sufficiently diverse.

This assumption holds in many practical scenarios. For instance, this assumption holds when the context is drawn i.i.d., and for any context  $x$, there is a  strictly positive probability that any action $a$ is selected in each round. Specifically, it there exists $\epsilon>0$ such that $\Pr(a\text{ is selected}|X)>\epsilon$, the assumption is satisfied.

\section{Additional Experimental Results}
\label{app:addexp}

\begin{figure}
    \centering
    \includegraphics[width=0.5\linewidth]{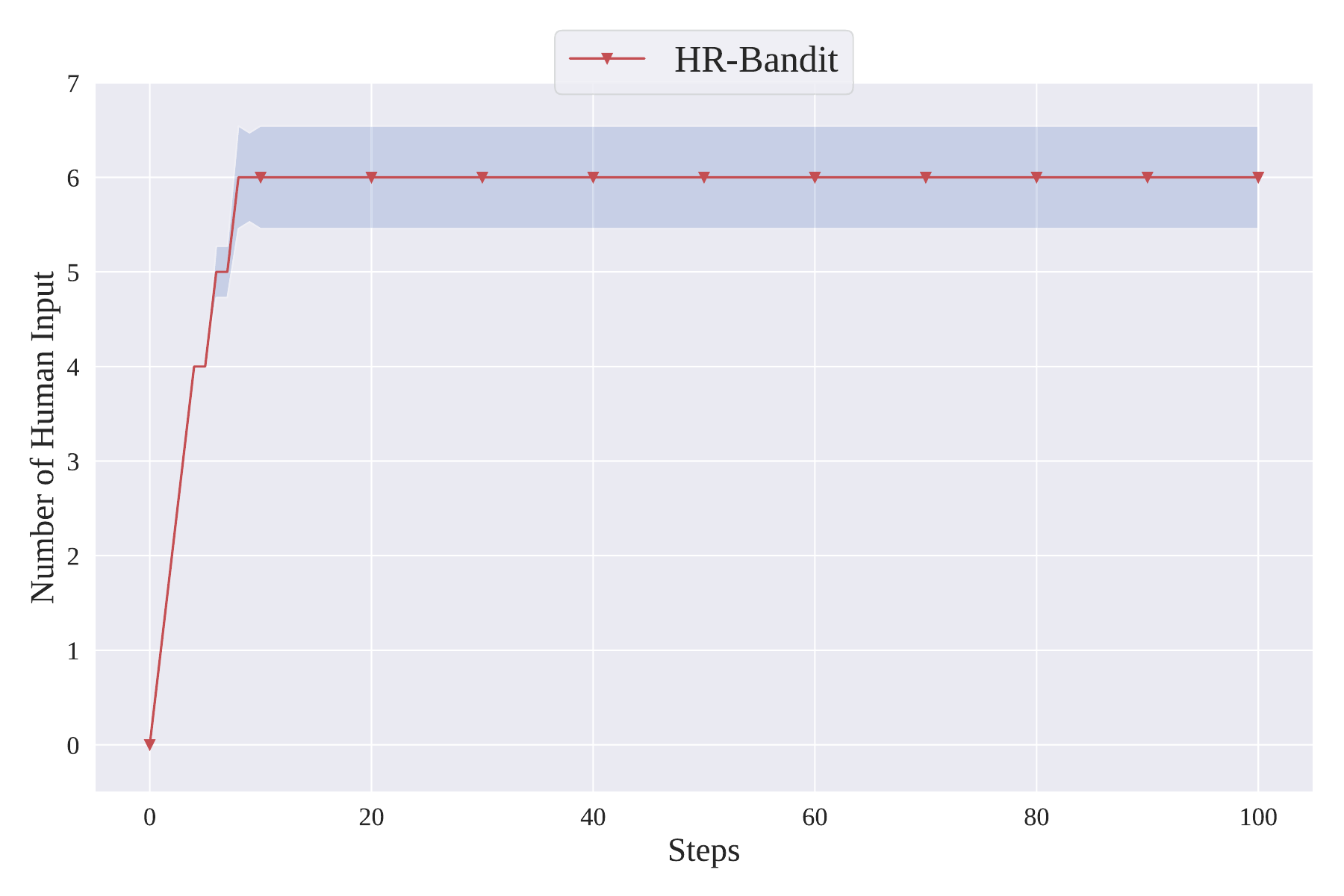}
    \caption{Number of Human Queries}
    \label{fig:ask_fert}
\end{figure}

The result on the number of human queries is shown in \Cref{fig:ask_fert}. Similar to our results in the main paper and the theoretical guarantee in \Cref{thm: human effort}, \textsf{HR-Bandit} only requires a limited number of human queries to warm-start the bandit algorithm and achieves a significantly better regret. 

In addition, we conduct experiments on the Infant Health and Development Program (IHDP) dataset \citep{hill2011bayesian}. IHDP dataset is used to accesss the effect of home visit from specialist doctors on the cognitive test scores of premature infants. The dataset has 25 features on infant and parents characteristics, we pick the first  5 features for simplicity and assume they are all mutable. The treatment is the home visit from specialist doctors and the outcome is the cognitive score. We fit a linear regression model on each treatment arm and use it as the ground-truth model to assess outcomes of the recourses not in the original dataset. 

\begin{figure}[h]
  \centering
  \begin{subfigure}[b]{0.45\textwidth}
    \includegraphics[width=\textwidth]{aistats25/fig/regret_syn.pdf}
    \caption{Recourse Regret}
    \label{fig:ihdp_regret}
  \end{subfigure}
  \hfill 
  \begin{subfigure}[b]{0.45\textwidth}
    \includegraphics[width=\textwidth]{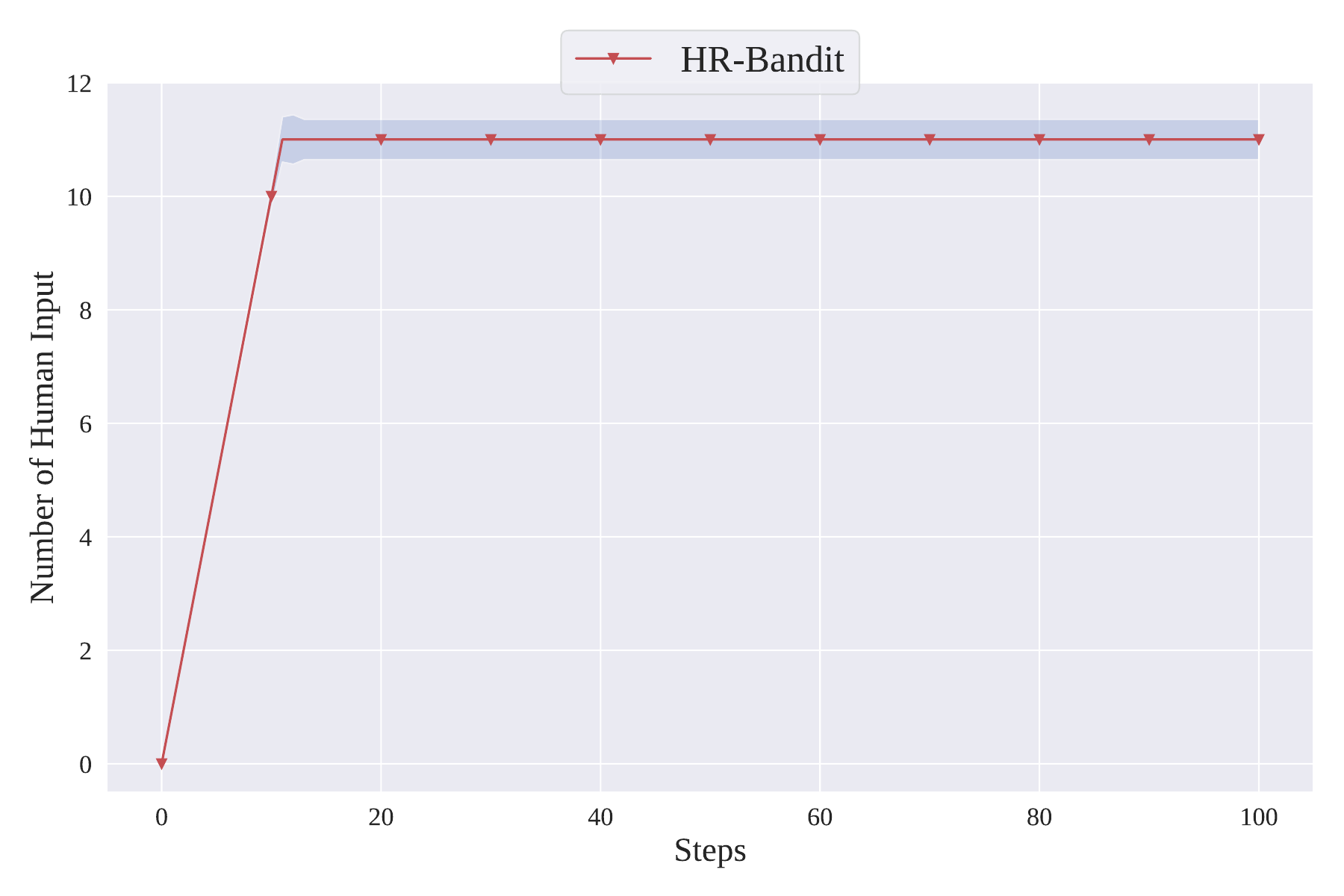}
    \caption{Human Queries}
    \label{fig:ihdp_human}
  \end{subfigure}
  \caption{IHDP Data.}
  \label{fig:ihdp}
\end{figure}

The results of the recourse regret and the number of human queries are shown in \Cref{fig:ihdp_regret} and \Cref{fig:ihdp_human}, respectively. Similar to our findings in the main paper, we observe \textsf{RLinUCB} achieves a sublinear regret (\Cref{thm: regret}), \textsf{HR-Bandit} achieves a significantly better (\Cref{thm: warm start multiple actions},\Cref{thm: improvementguarantee}) and sublinear regret (\Cref{thm: robustguarantee}) with only a finite number of human queries (\Cref{thm: human effort}). 

\end{document}